WORKSHOP ON

# Robotics & AI in Medicine

RESEARCH & TECHNOLOGY

## Final Report

---

*Dr. Juan P Wachs*
*Edwardson School of Industrial Engineering*
*Purdue University*
*315 N. Grant Street*
*West Lafayette, IN 47907*

*March 16, 2026*

## Table of Contents

## 1. Executive Summary:

The CARE Workshop on Robotics and AI in Medicine, held on December 1, 2025 in Indianapolis, convened leading researchers, clinicians, industry innovators, and federal stakeholders to shape a national vision for advancing robotics and artificial intelligence in healthcare [21]. The event highlighted the accelerating need for coordinated research efforts that bridge engineering innovation with real clinical priorities, emphasizing safety, reliability, and translational readiness with an emphasis on the use of robotics and AI to achieve this readiness goal.

Across keynotes, panels, and breakout sessions, participants underscored critical gaps in data availability, standardized evaluation methods, regulatory pathways, and workforce training that hinder the deployment of intelligent robotic systems in surgical, diagnostic, rehabilitative, and assistive contexts. Discussions emphasized the transformative potential of AI-enabled robotics to improve precision, reduce provider burden, expand access to specialized care, and enhance patient outcomes—particularly in underserved regions and high-risk procedural domains. Special attention was given to austere settings, disaster and relief and military settings.

The workshop demonstrated broad consensus on the urgency of establishing a national Center for AI and Robotic Excellence in medicine (CARE) [1]. Stakeholders identified priority research thrusts including human–robot collaboration, trustworthy autonomy, simulation and digital twins, multi-modal sensing, and ethical integration of generative AI into clinical workflows. Participants also articulated the need for high-quality datasets, shared testbeds, autonomous surgical systems, clinically grounded benchmarks, and sustained interdisciplinary training mechanisms.

Overall, the workshop provided compelling evidence of strong community demand and strategic alignment for launching the CARE initiative. The insights and recommendations generated form the foundation for a coordinated research agenda aimed at accelerating the safe, equitable, and scalable deployment of robotics and AI technologies in modern medicine.

## 2. Introduction and Need

The rapid convergence of robotics, artificial intelligence, and clinical medicine is reshaping the future of healthcare delivery. Yet, realizing the full potential of these technologies requires intentional coordination between engineering innovation and real-world clinical needs. To address this gap and chart a unified path forward, Purdue University’s Edwardson

School of Industrial Engineering, Weldon School of Biomedical Engineering, Purdue University, CTSI and the Indiana University School of Medicine jointly convened the CARE Workshop on Robotics and AI in Medicine on December 1, 2025 in Indianapolis.

This workshop served as a pivotal milestone in the formation of a new center, the Center for AI and Robotic Excellence in medicine (CARE)—a collaborative initiative designed to bridge disciplinary boundaries and accelerate translational research. Bringing together surgeons, clinicians, engineers, computer scientists, industry leaders, and federal research stakeholders, the event established a shared understanding of the challenges and opportunities surrounding intelligent robotic systems in healthcare.

The CARE Workshop underscored the pressing need for systematic research advances in trustworthy autonomy, surgical and interventional robotics, simulation and digital twins, multimodal sensing, and data-driven clinical decision support. Equally important, the discussions illuminated structural barriers—such as limited access to clinical datasets, lack of standardized evaluation frameworks, and accessible laboratories and diagnostic platforms —that currently slow innovation and hinder safe adoption.

By providing a platform for candid dialogue, cross-sector alignment, through keynotes, brainstorming sessions and panels, this forum helped in the identification of strategic research priorities. The workshop laid the intellectual and organizational groundwork for launching the new CARE Center in Indianapolis. This report summarizes the workshop's findings, synthesizes key insights from participants, and articulates a forward-looking agenda for a center that will position Indiana—and the nation—at the forefront of robotics and AI in medicine.

## The Challenge and Potential Solutions

There is a need to create necessary fundamental science, research and application tools to augment capabilities of surgeons and clinicians. Many problems plaguing clinical care today requires developing super intelligent robotic assistants capable of reasoning [2] (e.g. making complex decisions in real-time, minimizing human error and improving outcomes) and action (e.g. drawing blood for lab analysis, take vitals, operate in the OR and in constrained settings). These new capabilities will be tailored to deliver care appropriate to an individual's genetic makeup, pathology, and environment. When therapies cannot take place in-situ, telemedicine and telerobotics powered by super intelligent AI will offer treatment that is indistinguishable from in-person care [3]-[6]. Thus, healthcare access can be extended to remote areas experiencing health professional shortages and conflict zones. One way to achieve this goal is to have autonomous driving laboratories as they can overcome the structural barriers that limit access to timely, high-quality care. These mobile, self-navigating units can deliver point-of-care diagnostics, basic procedures, and telehealth services directly to patients, eliminating long travel times and dependence on scarce local providers. They enable rapid response during emergencies, support continuous monitoring for chronic conditions, and create resilient, distributed healthcare infrastructure. By combining autonomy, advanced sensing, and AI-driven clinical workflows, these laboratories bring modern medicine to regions historically left behind, dramatically improving equity, outcomes, and population-level health resilience [7].

As some of these developments, will require careful adjustment, evaluation and validation before actual deployment, simulation and digital twins are valuable tools. High-fidelity surgical simulation and robust, well-curated datasets are essential foundations for the future of AI-guided surgery and medicine. Simulators provide safe, controllable environments for training autonomous and semi-autonomous systems, enabling them to learn complex workflows, rare events, and edge-case scenarios without risk to patients. Such simulators are data thirsty and they are often as good as the data that they are trained with. Comprehensive datasets capturing anatomical variation, procedural nuances, and sensor-rich surgical data allow AI models to generalize reliably and support real-time decision-making [8]. Together, simulation and datasets accelerate algorithm development, improve system safety, and enable rigorous validation before clinical deployment. These resources will be central to building trustworthy, data-driven surgical technologies that enhance precision, reduce complications, and support surgeons across diverse clinical settings.

***How AI and Robotics Support Medicine:*** AI and robotics can enhance medicine and surgery by improving precision, reducing human error, and supporting clinicians with real-time decision guidance. They streamline workflows, enable minimally invasive procedures, expand access to expert care through teleoperation, and deliver consistent, data-driven performance—ultimately improving outcomes, safety, and efficiency across diverse clinical environments.

***How medicine can help Robotics & AI*:** Medicine and surgery can improve AI by providing rich, complex data and well-defined use cases that drive meaningful model development. Clinical challenges push AI to become more reliable, interpretable, and safe. Real-world surgical workflows also reveal edge cases and constraints, guiding AI systems toward higher precision, robustness, and true clinical relevance.

***The National Hub for AI, Robotics and Medicine:*** There is a pressing need to establish the Center for AI and Robotic Excellence in Medicine (CARE) in Indianapolis to unite Purdue University's engineering leadership with the Indiana University School of Medicine's clinical expertise. As AI and robotics rapidly transform healthcare, the region lacks an integrated hub capable of driving translational research, validating technologies in real clinical environments, and training the next-generation workforce. CARE would bridge engineering innovation with clinical needs, accelerate development of trustworthy autonomous systems, and create shared datasets, testbeds, and simulation platforms. Positioned between two world-class institutions, the center would advance patient outcomes, strengthen Indiana's biomedical ecosystem, and shape the future of intelligent healthcare.

## 3. Structure of the Workshop

### Opening Remarks

The workshop opened with welcoming comments from leadership at Purdue University and the Indiana University School of Medicine. Dr. Juan Wachs, the inaugural director of the CARE center, open the workshop, followed by remarks from Dr. Joseph Wallace, Associate Vice President for Research Development, Professor of Biomedical Engineering, Dr. Son Young-Jun Son, James J. Solberg Head and Ransburg Professor of Edwardson School of Industrial Engineering, and representing IUSM, Dr. Karl Bilimoria, who is the s the Chair of the Department of Surgery and the Jay Grosfeld Professor of Surgery at Indiana University School of Medicine. The Speakers emphasized the growing importance of uniting engineering innovation with clinical practice and highlighted the workshop's purpose: to identify strategic priorities for launching the Center for AI and Robotic Excellence in Medicine (CARE). The remarks set the tone for a collaborative, forward-looking event.

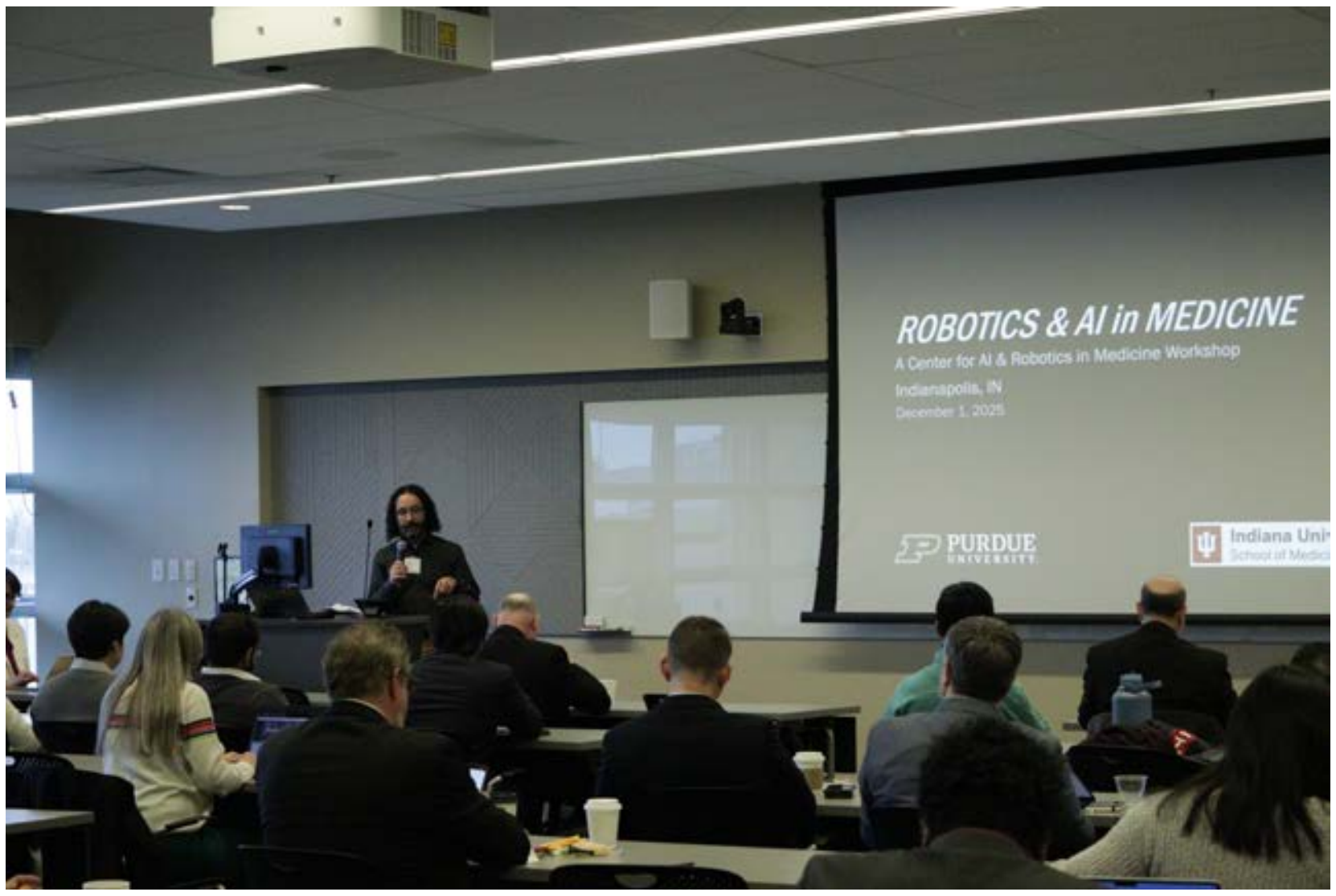

*Figure 1. Introduction to the Workshop by Dr. Joseph Wallace. Associate Vice President for Research Development, Professor of Biomedical Engineering*

### Keynotes

Keynote presentations delivered deep insights into the state of AI-driven robotics in surgery, diagnostics, and health systems. Clinical and engineering experts outlined current limitations, future opportunities, and the need for trustworthy autonomy. They also emphasized the urgency of building shared datasets, simulation resources, and translational pathways. The first keynote "Robotics and AI: Opportunities and Challenges" was given by Dr. Richard Satava (Washington University), who is a pioneer in surgical robotics and simulation, delivered a keynote highlighting the transformative potential of AI and autonomous systems in medicine, emphasizing innovation, safety, and precision as key drivers for the future of surgery and healthcare. The second keynote, was given by Dr. Jason Corso (University of Michigan) – "AI's Role in Upscaling Medical Practice". This talk highlights

AI’s potential to upskill healthcare, improving training and care. It explores visual AI in cardiothoracic surgery and interactive, physically-grounded guidance for practitioners, particularly enhancing access and quality in rural, underserved settings.

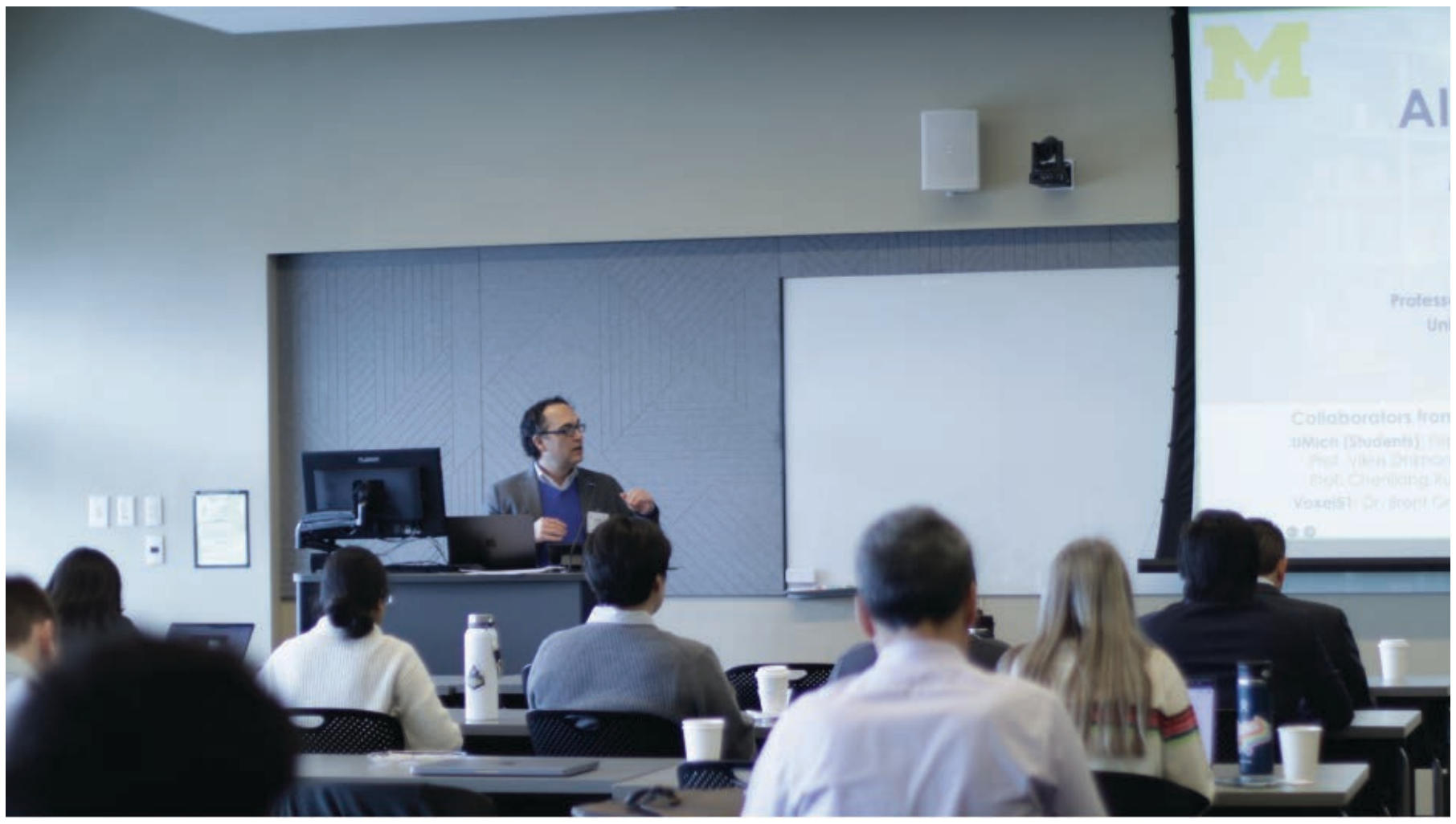

*Figure 2. Keynote from Prof. Jason Corso on “AI’s Role in Upscaling Medical Practice”.*

## Industrial Perspective

The Industry Lightning Presentations at the CARE Workshop featured brief, high-impact talks from leading companies shaping medical robotics and AI. Speakers included Dr. Tony Romano, Associate Director, (Zimmer Biomet), who discussed surgical robotics for transforming the future of total knee arthroplasty (TKA); Dr. Gordon G. Wisbach, MD, MBA, CAPT, MC, USN (RET) (Intuitive Surgical), presenting Intuitive Surgical’s latest innovations integrating AI-driven capabilities across the surgical continuum—from preoperative planning and intraoperative guidance to postoperative analytics.; Dr. Timothy Kowalewski, Associate Professor, University of Minnesota; CTO & Co-Founder, (LightSide Surgical), who talked about AI-driven surgical guidance tools combining simulation, and telerobotics advances AI, robotics, and surgical simulation by combining computer vision, real-time guidance, and telerobotics to enable “critical care anywhere.” . The last speaker was Dr. Rachel Clipp, who is the Assistant Director of Medical Computing (Kitware, Inc.), who provided an overview of Kitware’s expertise in intelligence in the medical space across a range of technical areas including computational physiology, medical triage, image analysis and segmentation, surgical simulation, and cyber-physical systems. Each presentation showcased emerging technologies, commercialization strategies, and real-world clinical applications, providing participants with insights into industry innovation, collaboration opportunities, and the translational challenges of deploying robotics and AI in diverse healthcare settings.

## Federal Perspective

Representatives from federal agencies provided insights on national priorities, funding directions, and the strategic relevance of centers like CARE. They discussed opportunities for alignment with federal initiatives in AI, robotics, and health innovation. We have representation from 3 agencies: DARPA, ARPA-H and NSF. Col. Jeremy Pamplin, MD - Program Manager, Biological Technologies Office, DARPA opened the session discussing key programs involving medical robotics and AI, such as the DARPA TRIAGE Challenge [9][9] and the DARPA PTG programs [10], and briefly the MASH programs [11]. Dr. Tyler Best, PhD - Acting Director, Health Science Futures, ARPA-H, talked about The Autonomous Interventions and Robotics (AIR) program [12] launched by ARPA-H which aims to catalyze the development of autonomous surgical robots. Last, Dr. Shivani Sharma, PhD - Program Director, NSF, discussed funding for medical research in the intersection of computing and AI. She presented programs such as SCH [13] and FRR [14]. After that each program director introduced their programs, a moderator asked questions about: 1. Funding Priorities & Vision; 2. Difference between Agencies (NSF vs DARPA vs ARPA-H); 3. Data, Standards & Interoperability; 4. Translation & Clinical Adoption & Success of a program; and 5 Regulation, Safety & Trustworthiness. Each program director presented their perspectives and the audience had the chance to ask questions and interact with the directors during this session and during the lunch break.

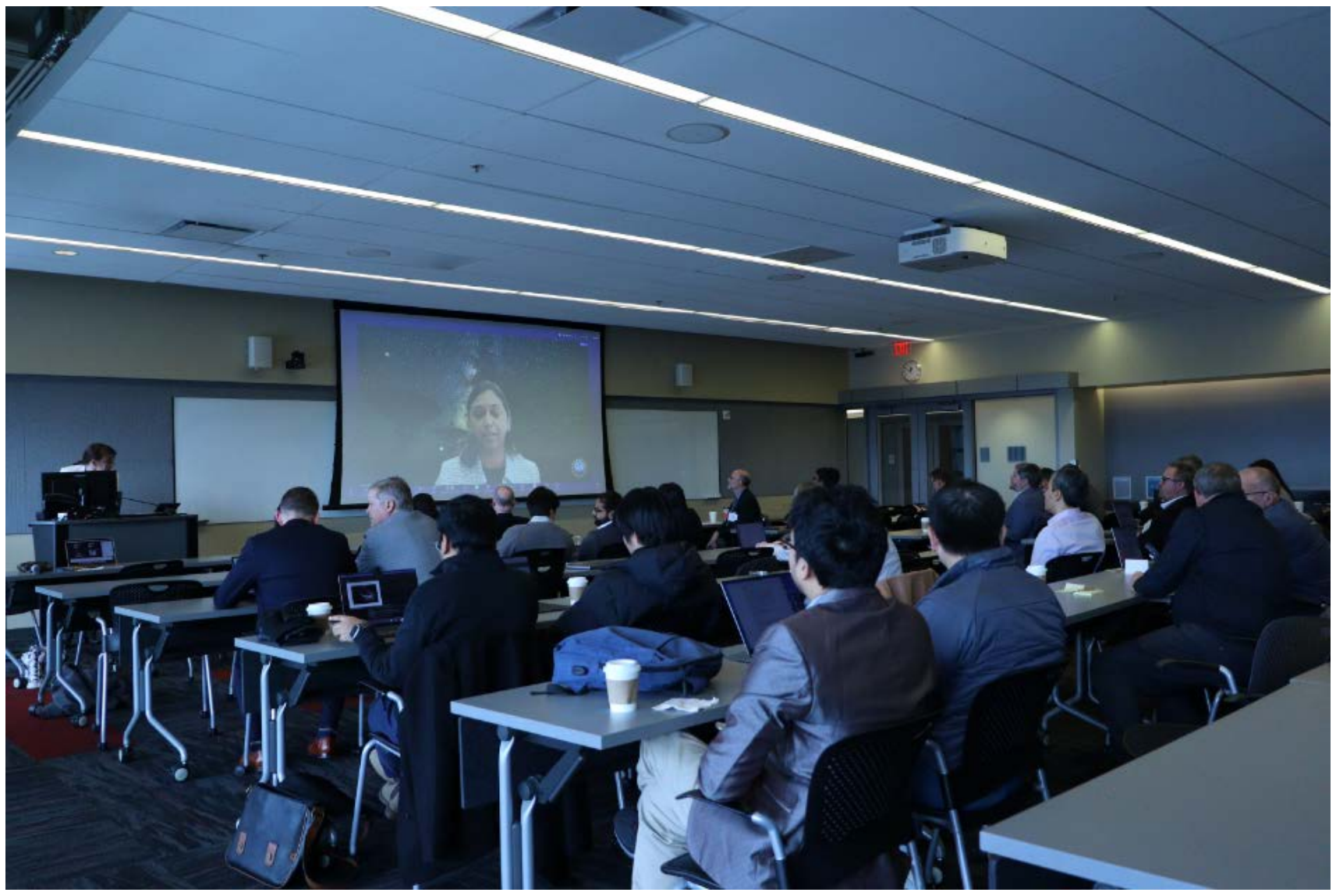

*Figure 3. Federal Perspective Session: Dr. Shivani Sharma about NSF programs supporting AI and medicine*

## Lighting Academic Talks

The program included four short talks, 15 minutes each, from faculty at Purdue and IUSM. From Purdue, Professors Denny Yu and David Cappelleri, from the Edwardson School of Industrial Engineering and the Weldon School of Biomedical Engineering, respectively, provided presentations. From IU School of Medicine, Professors Dimitrios Stefanidis and Andrew Gonzales from the Department of Surgery presented talks. Associate Professor

Denny Yu, presented the work entitled “Biobehavioral Sensing for Human-Aware Systems in Medicine” discussing the use of computer vision and physiological monitoring for objective understanding of human cognitive states during robotic surgery. Professor Dimitrios Stefanidis presented his work entitled “Artificial Intelligence in Surgical Education: Transforming Training, Assessment, and Skill Development” discussing the use of AI, such as large language models, computer vision, and deep learning, to augment surgical education by enabling objective skill assessment. Professor David Capelleri discussed his research entitled “AI Opportunities in Micro-robotic Surgery” about the development and use of microrobots for surgical applications, providing examples of different types of robots, for real-time surgeries and post-procedure analysis, highlighting where AI can enhance micro-robotic surgical applications. Last, Professor Andrew Gonzalez presented his work entitled “Labels, ontologies, and tasks: Oh My!!”, which discussed best practices for labelling, ontology development, and task formulation in the setting of interdisciplinary teams in the surgical setting. Over all, the discussions generated questions and open possibilities for further collaborations in research and development activities encompassing cross disciplinary and inter-institutional collaborations.

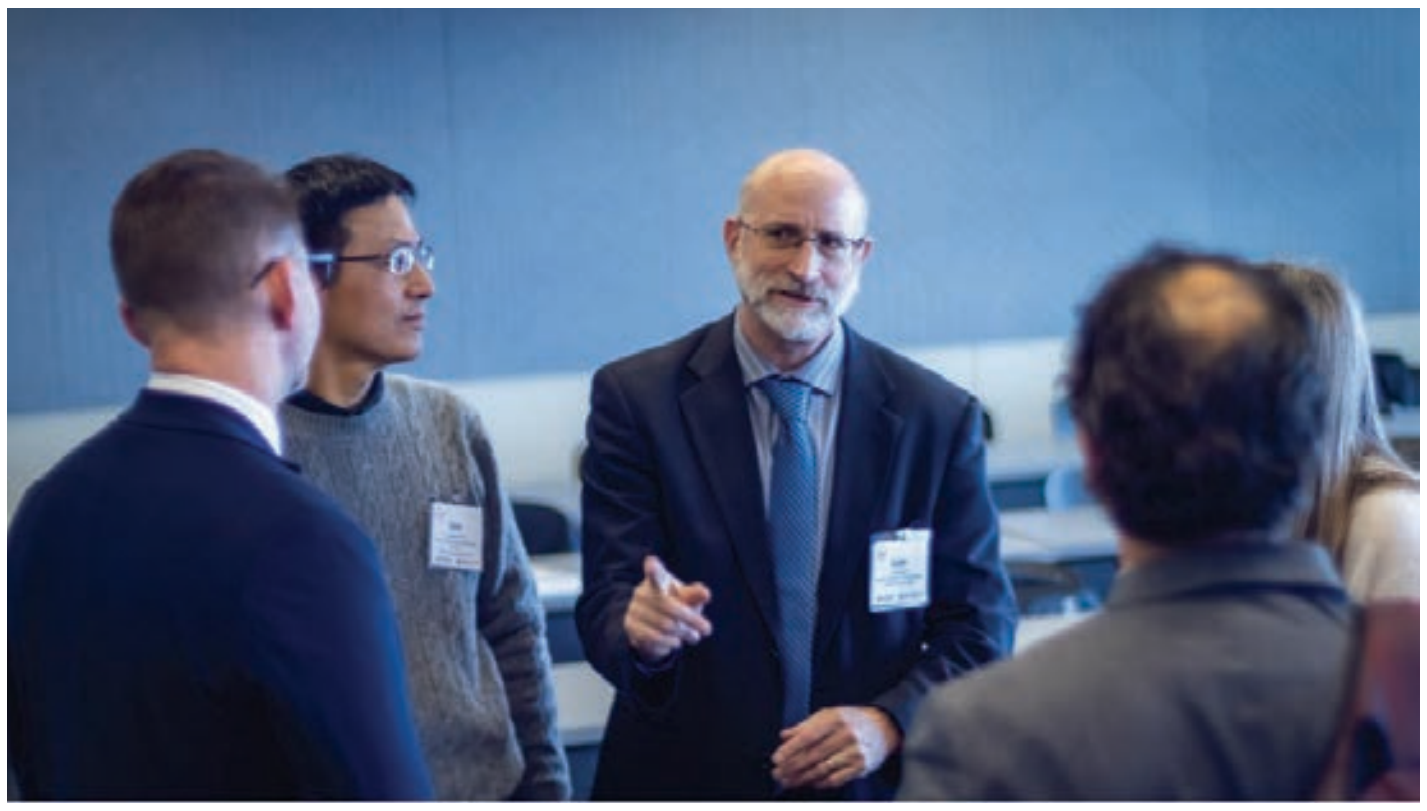

*Figure 4. Discussions with the program managers and organizers*

### Breakout Discussions

Participants engaged in focused breakout groups exploring data infrastructure, simulation, workflow integration, workforce training, and ethics. These discussions generated concrete recommendations shaping the vision and roadmap for the CARE Center. The discussions were structured around 3 themes: “Embodied AI”, “Field and Frontier Medicine” and “Autonomous Smart Laboratories”. An additional topic was added based on feedback from the participants involving “Datasets and Simulation”. These topics were selected based on the priorities of several research agencies, industrial partners and assessment of future development of R&D in the field of Med-Tech.

## 4. Breakout Session, Presentations and Discussions

Discussions were conducted by theme and based on preliminary assignments of the participants based on their preferences provided in a pre-attendance survey. The different groups gathered in rooms assigned each one for a different topic. Once each group was assembled, a led speaker introduced the main theme through 5-10 slides as a motivation kickstart step, and right after a set of questions were presented to each team. Overall, 30 minutes was assigned to all the teams to discuss the questions.

## The Questions

Overall, six questions were presented to each group to motivate the discussions and trigger suggestions and insights.

<u>Theme 1 : Embodied AI</u>

This questionnaire for this theme explores the potential role of autonomy and artificial intelligence in robotic surgery, focusing on clinical opportunities, technical feasibility, and future research directions. It seeks to identify which surgical procedures are most suitable for autonomous robotics based on factors such as cost efficiency, procedural complexity, surgeon workload, and case volume. It also examines “low-hanging fruit” opportunities—procedures not currently performed robotically but that may be well-positioned for early adoption. The questionnaire further encouraged the team to investigate the key challenges of integrating AI into robot-assisted surgery [15], including technological limitations, data requirements, safety concerns, regulatory barriers, and current system shortcomings. It asks which components of surgical procedures are most amenable to short-term versus long-term autonomy, and how varying levels of shared control between surgeon and machine could be implemented. Additionally, it calls for identifying three priority areas for research and development in autonomous surgical systems. Finally, it considers the potential role of humanoid robots in the operating room [16][17], exploring whether and how human-like robotic systems might contribute to surgical workflows.

<u>Theme 2 : Field and Frontier Medicine</u>

This questionnaire examines the future of surgical care and robotic collaboration in austere, remote, and battlefield environments. It explores how surgical mentoring—particularly telementoring [18] —may evolve to support remote procedures in resource-limited or high-risk settings. A key focus is the interaction between medics and robotic systems in first-response or combat medicine, including how responsibilities might be divided and which procedures could be performed collaboratively and by afar. The questionnaire also investigates the technical, logistical, and ethical challenges of deploying fully autonomous surgical agents in field environments, where infrastructure, connectivity, sterility, and stability may be limited. It considers optimal methods for delivering real-time instruction and decision support during remote surgeries, especially when expert surgeons are not physically present. Additionally, it evaluates the role of AI in guiding, mentoring, and assisting medics in the field, including decision-making support and procedural coaching. Finally, it addresses the need for flexibility and improvisation (referred as to “just in time”) in field surgery, questioning how AI systems could accommodate deviations from standard protocols while maintaining safety and effectiveness.

Theme 3 : Autonomous Smart Laboratories

This questionnaire explores the concept and applications of Autonomous Driving Laboratories (ADLs) in medicine and biomedical research. It seeks to identify existing examples of ADLs and evaluate their relevance to healthcare and experimental science. The questions examine which types of biological experiments—across patients, animal models, and microorganisms—are most suitable for automation, including considerations of scalability and the value of high-throughput experimentation. The questionnaire also considers whether ADLs could extend beyond research to support diagnosis and treatment in rural or underserved settings [19], highlighting associated technical, ethical, and logistical challenges. It addresses the operational and safety requirements for handling diverse biological systems, as well as regulatory considerations, particularly when working with fragile or sensitive species.

Finally, it explores the role of AI in enhancing ADLs, including experimental design, data analysis, process optimization, and decision-making support within automated laboratory environments.

Theme 4 : Datasets and Simulation

This questionnaire examines the role of computational simulation in training and developing surgical robotic systems and AI-driven medical technologies. It begins by exploring how multiscale biological dynamics—ranging from cellular processes to organ-level physiology—can be modeled in patient simulations [20]. It then considers whether high-fidelity computer simulations are necessary for training surgical robots and how such simulations contribute to skill acquisition and system validation. The questionnaire also investigates the potential of gamification and reinforcement learning as effective strategies for robot training, questioning whether immersive or game-like environments can accelerate learning. It addresses the importance of realism in simulations, particularly how closely virtual environments should replicate human anatomy and physiology to ensure effective transfer to real-world settings.

Finally, it focuses on strategies to bridge the sim-to-real gap, identifying best practices to ensure that behaviors learned in simulation generalize safely and reliably to clinical environments. Last, it explores how simulation can support AI development through data augmentation, synthetic data generation, and scenario diversification to enhance robustness and performance.

## Brainstorming Areas and Main Findings

### Theme 1 : Embodied AI

*Summary of Responses on Autonomous Robotics and AI in Surgery*

The responses outline a comprehensive vision for the evolution of autonomous robotics and AI in surgery, identifying where autonomy is most feasible, what challenges remain, and which research directions will shape the future operating room.

*1. Surgeries Most Likely to Benefit from Autonomy*

Autonomous robotics is expected to advance first in procedures that are highly standardized, imaging-guided, repetitive, and geometrically predictable. Orthopedics (e.g., joint replacements, spinal instrumentation) and neurosurgery (e.g., stereotactic procedures) are prime candidates due to rigid anatomy and precise preoperative planning. Catheter-based interventions, needle-based biopsies, and certain laparoscopic subtasks (suturing, camera control) are also strong early applications. Autonomy thrives in structured environments where variability is limited and precision is paramount, while procedures involving deformable tissue, bleeding, and complex intraoperative decision-making will adopt autonomy more gradually.

*2. Low-Hanging Fruit for Robotic Expansion*

Several high-volume procedures remain under-roboticized despite being well-suited for automation. These include breast-conserving surgery, thyroid and parathyroid surgery, hernia repairs, ENT procedures (sinus and ear surgery), vascular access, and image-guided injections or biopsies. These operations share characteristics such as predictable anatomy, repetitive workflows, and imaging guidance, making them natural candidates for semi-autonomous robotic assistance. Expanding robotics into these areas could significantly improve precision, standardization, and outcomes.

*3. Challenges of Integrating AI into Robotic Surgery*

Major barriers to AI integration include:

- **Soft tissue unpredictability**, deformability, and dynamic surgical environments.
- **Limited high-quality annotated datasets** and fragmented data infrastructure.
- **Safety, reliability, and regulatory constraints**, given the high stakes of surgical error.
- **Proprietary robotic platforms** that limit AI integration and experimentation.
- **Cultural and economic barriers**, including surgeon trust and high implementation costs.

While technologies such as advanced computer vision, multimodal sensing, reinforcement learning simulators, and semi-autonomous robotic systems already exist, they remain fragmented and limited to narrow tasks. True holistic autonomy requires significant advances in perception, modeling, and system integration.

*4. Short-Term vs Long-Term Autonomy and Shared Control*

Autonomy will emerge incrementally at the task level:

- **Short-term autonomy**: camera control, suturing along predefined paths, bone cutting, needle placement, and other geometric or constrained tasks.
- **Long-term autonomy**: soft tissue dissection, bleeding control, adaptive intraoperative decision-making, and complex anastomoses—tasks requiring contextual reasoning and adaptability.

Shared control represents the most realistic near-term model, combining robotic precision with human oversight. Robots may execute predefined subtasks while surgeons supervise, intervene, or adjust strategy. This layered autonomy approach enhances safety and collaboration rather than replacing surgeons.

*5. Three Key Research and Development Areas*

Three foundational research domains are highlighted:

1. **Real-time multimodal perception and soft tissue modeling**
   Integration of video, force feedback, imaging, and predictive tissue models to enable safer adaptive autonomy.
2. **Task- and procedure-level policy learning**
   Advancing reinforcement learning, imitation learning, physics-aware digital twins, and high-fidelity simulators to teach robots adaptive surgical strategies.
3. **Human–robot shared control and explainable autonomy**
   Developing intuitive interfaces, uncertainty modeling, dynamic control transfer, and transparent AI reasoning to support safe collaboration and regulatory approval.

These areas collectively support the transition from teleoperated tools to embodied intelligent surgical systems.

*6. Role of Humanoid Robotics in the Operating Room*

Humanoid robots are envisioned not as primary surgical operators but as intelligent logistical and environmental assistants within the operating room. Because ORs are designed for human morphology, humanoid systems could:

- Manage perioperative logistics and equipment positioning

- Assist in patient handling and positioning
- Support OR turnover and sterile setup
- Facilitate coordination between multiple robotic systems

Their role would be infrastructural and collaborative, enhancing efficiency and enabling specialized surgical robots to operate more autonomously. Humanoid robots may also serve as research platforms for multi-agent embodied AI and human–robot collaboration.

*Conclusions*

The responses collectively describe a future in which surgical autonomy evolves gradually through structured task automation, physics-aware learning, and collaborative control models. Early adoption will occur in predictable, image-guided procedures, while complex adaptive surgery remains human-led for the foreseeable future. The long-term trajectory is not surgeon replacement, but the emergence of intelligent, perceptive robotic partners integrated into a coordinated surgical ecosystem.

## Theme 2: Field and Frontier Medicine

*Summary of Responses on Field and Frontier Medicine:*

The responses outline a forward-looking vision for surgical care in austere, remote, and battlefield environments, emphasizing human–robot collaboration, adaptive tele-mentoring, and the significant technical and ethical barriers to full autonomy.

*1. The Future of Tele-Mentoring in Austere Settings*

Future surgical tele-mentoring is expected to evolve toward a **semi-autonomous, human-in-the-loop model** rather than full robotic autonomy. In this framework, autonomous systems perform structured subtasks while remote experts retain supervisory control over critical decisions.

A key design principle is the distinction between **crisis and non-crisis modes**.

- In non-crisis situations, systems can support bidirectional dialogue, detailed explanations, and educational mentorship.
- In crisis scenarios, the system must prioritize speed and clarity, suppressing non-essential information to deliver concise, life-saving instructions.

Tele-mentoring platforms must also be **context-aware**, adapting guidance based on available resources, infrastructure, and geopolitical constraints. The operational meaning of “Golden Hour” care varies significantly between rural healthcare systems and resource-limited combat zones, requiring dynamic objective framing (e.g., maximizing survival vs. preserving long-term function).

Looking further ahead, the discussion envisions transformative biomedical technologies—such as metabolic suppression (synthetic torpor), neuro-modulation, and bioprinting—that could redefine trauma stabilization and triage. In such a future, tele-mentoring might extend beyond immediate repair toward biological preservation and regeneration strategies.

*2. Human–Robot Teaming in First Response and Battlefield Medicine*

In high-stress environments, medics face cognitive overload and bias. Robotic systems can serve as **objective validators**, using multimodal sensing and machine vision to cross-check vital signs, suggest alternative diagnoses, and monitor procedural fidelity (e.g., AED pad placement).

In signal-denied settings where real-time telecommunication is unavailable, robots may shift from communication platforms to **autonomous instructional systems**. Instead of static reference libraries, they would curate dynamic, step-specific video modules and project augmented reality overlays directly onto the patient for just-in-time guidance.

Trust is central to effective collaboration. Future systems may incorporate **sentiment recognition**, adjusting communication style to match medic stress levels, simplifying language, and reinforcing confidence during invasive procedures.

Scalability in mass-casualty scenarios may rely on a **1-to-N command hierarchy**, where a single remote expert directs multiple semi-autonomous robotic sub-agents. The expert provides high-level intent, while robots execute low-level tasks autonomously, maximizing limited expert cognitive capacity.

*3. Challenges of Fully Autonomous Surgery in Austere Settings*

Several major obstacles limit the feasibility of complete autonomy in field environments:

- **Lack of robust, high-quality datasets** reflecting the chaotic and heterogeneous nature of trauma care. Field medicine is high-entropy and unpredictable, making generalization difficult.
- **Limited explainability and trust** in AI decision-making, especially given risks of hallucination or misidentification of anatomical structures.
- **Asymmetrical error tolerance**: society accepts human error under stress but expects near-perfection from robots. Even statistically superior robotic performance may be rejected after a single visible failure.

- **Ethical constraints**, particularly the prohibition against systems "learning on the fly" in life-threatening contexts, which conflicts with iterative machine learning paradigms.
- **Absence of standardized benchmarking frameworks** to define and measure machine surgical competence. Without clear metrics and regulatory standards, certification and deployment remain impractical.

*Summary*

The discussion presents a future in which autonomous systems enhance—but do not replace—human expertise in frontier medicine. Near-term progress will focus on semi-autonomous assistance, adaptive tele-mentoring, and intelligent instructional support. Fully autonomous field surgery remains constrained by data limitations, ethical considerations, trust deficits, and regulatory gaps.

Ultimately, the path forward lies in building context-aware, trustworthy, and hierarchically coordinated human–robot systems capable of operating effectively in the uncertainty and resource scarcity that define austere medical environments.

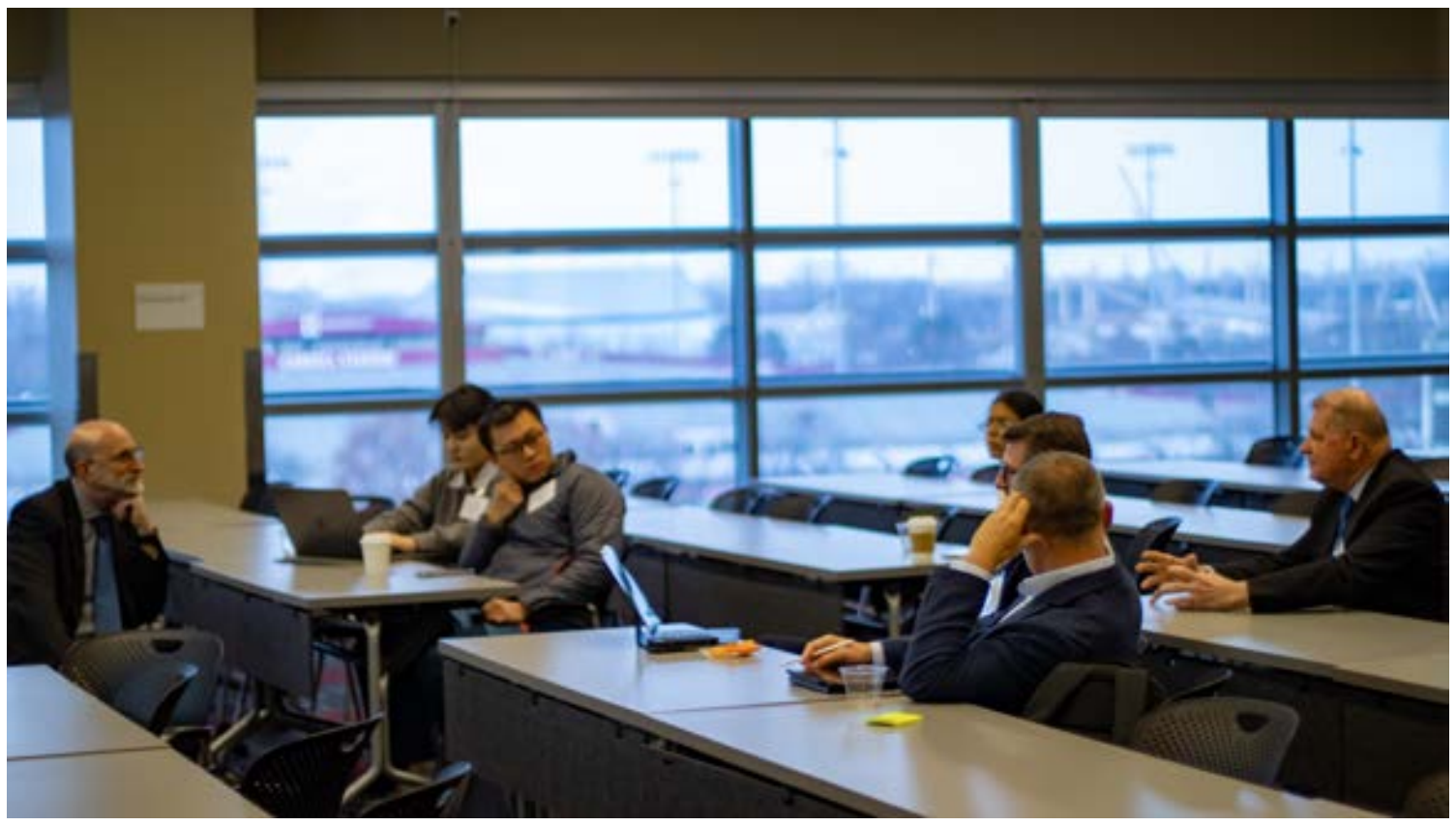

*Figure 5. Breakout Discussion Sessions with students, industrial partners, academics and practitioners*

## Theme 3: Autonomous Driving Laboratories (ADLs)

*Summary of Responses on Autonomous Driving Laboratories (ADLs)*

The discussion explores the emerging role of Autonomous Driving Laboratories (ADLs) in healthcare, highlighting current applications, experimental opportunities, and their potential impact—particularly in rural and mobile care settings.

*1. Current and Emerging Examples of ADL in Medicine*

Existing examples of ADL applications include **autonomous drug delivery systems** (e.g., pilot programs in New Zealand) and **physical rehabilitation platforms**, where automation improves consistency and reduces human variability. A central theme was that reducing human intervention can lower management complexity, minimize errors, and increase repeatability.

Beyond these, participants identified expanding applications of autonomous systems in healthcare logistics and operations. These include:

- Autonomous transport of medical supplies and laboratory specimens within hospital campuses
- UV-based robotic disinfection systems to enhance infection control
- Autonomous navigation within surgical robotics to improve procedural precision
- Driverless patient transport shuttles to improve mobility and access to care

Collectively, these examples demonstrate how autonomous mobility and laboratory systems can improve efficiency, safety, and patient experience while reducing human workload.

*2. Suitable Experiments and High-Throughput Applications*

Numerous experimental and diagnostic activities were identified as suitable for ADL deployment, particularly those that are structured, repetitive, and protocol-driven. These include:

- Imaging procedures
- Clinical trials
- Biopsies and biopsy analysis under defined conditions
- Routine non-invasive check-ups
- Telemedicine and video-based consultations
- Various standardized diagnostic procedures

The concept of a **mobile ADL platform**—essentially a self-contained, autonomous medical vehicle—was strongly supported. Such systems could provide telemedicine services and diagnostic capabilities directly to communities, reducing reliance on traditional brick-and-mortar facilities and minimizing patient travel.

High-throughput capability was viewed as advantageous for standardized diagnostics and screening programs, where repeatability and automation can improve efficiency and scalability. However, concerns were raised about increased biomedical and material waste generation, and the logistical challenges of managing that waste responsibly in mobile settings.

*3. ADL for Rural Diagnosis and Treatment: Opportunities and Challenges*

There was strong support for deploying ADLs in rural and underserved areas, particularly for **early triage, diagnostics, and preliminary treatment**. Mobile autonomous platforms could extend access to healthcare where infrastructure and specialist availability are limited.

However, several operational and systemic challenges were identified:

- Maintaining cleanliness and sterilization within mobile units
- Staffing, training, and oversight requirements
- Reliable supply chain management for medical goods
- Mechanical maintenance and physical upkeep of vehicles
- Physical accessibility in remote or difficult terrain
- Patient awareness, trust, and acceptance of autonomous systems
- Ensuring consistent quality of care across locations

Uniform standards of care and quality assurance were emphasized as essential for safe implementation.

*Summary*

The responses reflect optimism about the role of ADLs in increasing healthcare accessibility, efficiency, and repeatability—particularly in mobile and rural contexts. While technological feasibility is advancing rapidly, successful implementation will depend on addressing infrastructure, waste management, regulatory standards, and patient trust. ADLs are seen not merely as autonomous vehicles, but as integrated mobile healthcare ecosystems capable of reshaping how and where medical services are delivered.

## Theme 4 : Datasets and Simulation

*Summary of Responses: Datasets and Simulation in Surgical Robotics*

The discussion highlights the growing importance of multiscale modeling, computer simulation, reinforcement learning, realism calibration, and sim-to-real transfer in advancing AI-driven surgical robotics. Across all questions, a central theme emerges: simulation is becoming foundational—not only for training robots, but for building safe, scalable, and data-efficient intelligent systems.

*1. Modeling Multiscale Dynamics in Patient Simulation*

Multiscale patient simulation requires modeling biological processes across multiple levels—molecular, cellular, tissue, organ, and whole-body systems. The key challenge lies not in modeling each level independently, but in **integrating them efficiently and meaningfully**.

Participants emphasized:

- Modeling only the **minimum scale necessary** to answer the specific problem.
- Using **graph-based or modular approaches** to define transition points ("off-ramps") where outputs from one scale become inputs to another.
- Balancing **fidelity with computational feasibility**, especially for real-time applications.
- Recognizing that higher granularity dramatically increases computational demands.

The consensus was that full-detail modeling is neither practical nor necessary. Instead, simulation should be purpose-driven, computationally efficient, and strategically layered. Multiscale modeling is increasingly important for AI training and next-generation medical simulation, but must be built around targeted objectives.

*2. Is Computer Simulation Necessary for Training Surgical Robots?*

While not yet universal, simulation is increasingly viewed as essential for surgical robot development.

Key arguments in favor include:

- **Cost efficiency** compared to physical prototypes.
- Ability to generate **rare and edge-case scenarios** not easily captured in clinical practice.
- Scalable data generation for AI training.
- Reduced reliance on human subjects.
- Growing **regulatory pressure (e.g., FDA)** toward simulation-based validation.

Compared to industries like aerospace and automotive automation, surgical robotics is considered behind in simulation adoption. Many participants suggested that the primary long-term value of simulation lies in **training AI systems**, rather than training human surgeons.

Challenges remain—high computational costs, realism limitations, and cultural resistance—but the overall trajectory points toward simulation becoming a standard component of surgical robotics development.

*3. Gamification and Reinforcement Learning (RL)*

Gamification was discussed from two perspectives: AI training and human engagement.

For AI systems:

- RL already uses metric-based “game” structures via reward functions.
- AI agents can compete or self-optimize based on defined utility functions.
- Similar strategies are used in industries like autonomous driving and humanoid robotics (e.g., Tesla, Nvidia).

For humans:

- Gamification may significantly enhance engagement among surgeons and trainees.
- Techniques like leaderboards, scoring systems, and visual feedback can motivate participation.

However, caution is required:

- Poorly designed reward systems may incentivize incorrect behaviors.
- Focus on “scoring” must not undermine actual skill acquisition.

The consensus was that gamification can be powerful, particularly for human engagement, but requires careful alignment between incentives and desired learning outcomes.

*4. How Realistic Should Simulations Be?*

A major theme was strategic realism.

Participants emphasized:

- Not every detail requires perfect replication.
- The “80/20 rule” applies—functional realism is often sufficient.
- Prioritize fidelity in elements critical to the task (e.g., tissue strain, boundary identification, deformation mechanics).
- Deprioritize less essential elements (e.g., exact visual coloring) unless needed for specific applications.

Simulation realism should be driven by purpose:

- Marketing demos require high visual realism.
- AI training requires functional and mechanical accuracy.
- Computer vision models do not require perfect aesthetic detail.

The recommendation was to start with a **minimum viable simulation** and incrementally increase complexity as needed. Computational constraints make indiscriminate realism impractical.

*5. Bridging the Simulation-to-Real Gap*

Bridging the sim-to-real gap requires a structured and incremental strategy:

*Key Best Practices*

*1. Targeted Progression*

- Begin with narrow, simple objectives.
- Expand gradually toward more complex behaviors.

*2. Variation and Edge Cases*

- Use simulation to generate diverse scenarios.
- Intentionally introduce variability to improve robustness.

*3. Transfer Learning*

- Identify similarities between procedures.
- Transfer knowledge across related tasks when possible.

*4. Synthetic Data and Augmentation*

- Generate morphed anatomical datasets.
- Use simulation to assist labeling and validation.

*5. Resource-Aware Design*

- Balance realism with computational and economic constraints.

*6. Interdisciplinary Learning*

- Borrow from aviation, gaming, entertainment, and other industries with advanced simulation ecosystems.

*7. Validation and Regulatory Alignment*

- Develop benchmarking frameworks.
- Demonstrate simulation reliability for regulatory compliance.

The overarching message was that sim-to-real transfer should be deliberate, iterative, and guided by practical clinical utility rather than theoretical completeness.

*Summary*

Across all sections, a coherent vision emerges; simulation and data-driven modeling are becoming foundational to the development of surgical robotics and AI systems. Multiscale modeling must be purposeful and efficient. Simulation is increasingly indispensable for safe, scalable AI training. Gamification and reinforcement learning require carefully aligned incentives. Realism should be strategic rather than absolute. And bridging the sim-to-real gap demands incremental validation and interdisciplinary collaboration.
Together, these principles form the groundwork for a future in which intelligent surgical systems are trained, tested, and refined within sophisticated digital environments before entering the operating room.

# 5. Conclusions and Main Findings

The CARE Workshop on Robotics and AI in Medicine revealed strong consensus that the convergence of robotics, artificial intelligence, and clinical medicine has reached an inflection point. Participants agreed that the technological foundations for intelligent robotic systems already exist, but coordinated national infrastructure, standardized evaluation, and translational pathways are urgently needed to safely scale deployment across healthcare environments. The workshop clarified both the extraordinary potential of AI-enabled robotics and the structural barriers that must be addressed to realize that potential.

## Autonomy in Medicine Will Be Incremental, Layered, and Human-Centered

Across discussions on embodied AI and surgical robotics, a clear trajectory emerged: autonomy will not replace clinicians, but rather augment them through layered, task-level integration. Early adoption will occur in structured, imaging-guided, and geometrically predictable procedures such as orthopedics, neurosurgery, catheter-based interventions, and needle-guided tasks. Complex adaptive surgery involving deformable tissue and dynamic complications will remain human-led for the foreseeable future.
Shared control models—where robots execute constrained subtasks under surgeon supervision—were identified as the most realistic near-term framework. Advancing toward trustworthy autonomy requires breakthroughs in multimodal perception, physics-aware modeling, adaptive policy learning, and explainable human–robot interaction. Participants emphasized that autonomy must be measurable, interpretable, and clinically benchmarked to gain regulatory and professional acceptance.

## Data Infrastructure and Simulation Are Foundational

A recurring finding across all breakout themes was that progress is fundamentally limited by fragmented datasets, insufficient annotation standards, and the absence of shared testbeds. High-quality, multimodal clinical datasets—capturing anatomy, kinematics, imaging, and procedural context—are essential to train, validate, and benchmark AI systems.
Simulation and digital twins were identified as indispensable tools for accelerating development while maintaining safety. Rather than pursuing maximal realism, participants advocated for purpose-driven, computationally efficient simulation environments that focus on functional fidelity. Simulation enables rare-event modeling, scalable data augmentation, and reinforcement learning in safe environments. However, bridging the sim-to-real gap requires structured validation frameworks, incremental deployment strategies, and interdisciplinary collaboration.
The workshop highlighted the need for nationally coordinated simulation platforms and shared digital ecosystems capable of supporting autonomous surgical systems, regulatory validation, and AI training at scale.

## Field and Frontier Medicine Represents Both Urgency and Opportunity

Discussions on austere environments, disaster response, and battlefield medicine underscored the importance of robotics and AI for expanding care beyond traditional hospital settings. Participants envisioned semi-autonomous tele-mentoring systems that adapt between crisis and non-crisis modes, providing context-aware guidance tailored to resource constraints.
Human–robot teaming in high-stress environments emerged as a critical research domain. Robots may serve as cognitive stabilizers—validating diagnoses, providing procedural fidelity checks, and delivering just-in-time instructional support through augmented interfaces. However, full autonomy in these settings faces major barriers, including limited trauma datasets, ethical constraints against real-time algorithmic learning, and a lack of standardized benchmarks for machine competence.
The group emphasized that scalable, trustworthy, hierarchical human–robot systems—rather than fully independent robotic surgeons—will define near-term progress in frontier medicine.

## Autonomous Smart Laboratories and Mobile Care Platforms Expand Access

The concept of Autonomous Driving Laboratories (ADLs) gained strong support as a mechanism for decentralizing care delivery. These mobile, self-navigating units could provide diagnostics, imaging, telemedicine, triage, and high-throughput screening directly to underserved communities, rural regions, and emergency zones.
Participants identified structured, protocol-driven tasks—such as imaging, biopsies, routine diagnostics, and clinical trial support—as well-suited for automation. However, successful deployment requires solutions for sterilization, supply chain management, waste handling, infrastructure resilience, and public trust.
ADLs were viewed not simply as vehicles, but as integrated mobile healthcare ecosystems capable of enhancing equity, resilience, and population-level health outcomes.

## Trust, Standards, and Regulatory Alignment Are Critical Gaps

A major cross-cutting finding was the asymmetry in societal expectations between human clinicians and autonomous systems. While human error under stress is tolerated, robotic systems are expected to approach perfection. This reality places extraordinary importance on explainability, benchmarking, safety guarantees, and regulatory clarity.
The absence of standardized definitions of "machine surgical competence" currently impedes certification and comparison across platforms. Participants called for the development of clinically grounded benchmarks, interoperable data standards, and shared validation frameworks aligned with federal agency priorities (NSF, DARPA, ARPA-H).
Trustworthiness—technical, ethical, and social—was repeatedly identified as the central prerequisite for widespread adoption.

## The Case for a National CARE Center

Perhaps the most significant conclusion of the workshop was the broad alignment around the need for a dedicated national hub—the Center for AI and Robotic Excellence in Medicine (CARE). Such a center would:

- Integrate engineering innovation with clinical expertise.
- Develop shared datasets, testbeds, and simulation platforms.
- Advance research in embodied AI, trustworthy autonomy, and human–robot collaboration.
- Provide translational pathways from laboratory to clinical deployment.
- Train the next-generation interdisciplinary workforce.
- Align with federal priorities and industry needs.

Positioned between Purdue University and Indiana University School of Medicine, CARE would create a unified infrastructure capable of accelerating safe, equitable, and scalable deployment of robotics and AI in medicine.

## Concluding Remarks

The CARE Workshop demonstrated that the community is ready—and urgently motivated—to move from isolated innovation to coordinated national strategy. Robotics and AI have the potential to enhance precision, reduce provider burden, expand access to care, and strengthen healthcare resilience across conventional, rural, and austere settings. However, achieving this future requires investment in data infrastructure, simulation ecosystems, standardized evaluation, and human-centered autonomy frameworks.

The main finding is clear: the next era of medicine will be defined not by replacing clinicians, but by building intelligent, trustworthy robotic partners embedded within a coordinated, data-driven healthcare ecosystem. Establishing CARE in Indianapolis represents a strategic and timely step toward realizing that vision.

## 6. Organizers are Participants:

**Organizers:**

Juan P Wachs, Purdue University
Karl Bilimoria, Indiana University School of Medicine
Julia Sibley, Purdue University

**Participants:**

| | |
|---|---|
| Tyler Best | ARPA-H |
| Tony Romano | Zimmer-Biomet |
| Paul Thienphrapa | ARPA-H |
| Yanni Pandelidis | Cook Medical |
| Jeremy Pamplin | DARPA |
| Andrew Isch | Cook Medical |
| Gabriel Gruionu | DRG Innovation |
| Gordon Wisbach | Intuitive Surgical |
| David A Peterson | IU Health Bloomington |
| Vicky Thompson | IU Health Bloomington |
| Andrew Gonzales | Indiana University School of Medicine |
| Karl Bilimoria | Indiana University School of Medicine |
| Matt Ritter | Indiana University School of Medicine |
| Thomas Hayward | Indiana University School of Medicine, Veteran's Affairs |
| Natasha Sanz | Indiana University School of Medicine |
| Ping Li | Indiana University School of Medicine |
| Xiaoling Zhong | Indiana University School of Medicine |
| Dimitrios Stefanidis | Indiana University School of Medicine |
| Rachel Clipp | Kitware |
| Jeff Berkley | Light Side Surgical |
| Timothy M Kowalewski | Light Side Surgical |
| Shivani Sharma | NSF SCH, BMMB |
| Rick Satava | Professor Emeritus of Surgery |
| Abhinaba Bhattacharjee | Purdue University |
| Alfredo Ocegueda | Purdue University |
| Daniel Kuratomi | Purdue University |
| Gloria Yutong Zhang | Purdue University |
| Md Masudur Rahman | Purdue University |
| Mike Wozniak | Purdue University |
| Pronoma Banerjee | Purdue University |
| Shubh Mehta | Purdue University |
| Yupeng Zhuo | Purdue University |
| Zhixian Hu | Purdue University |
| Mike Foo CheBin | Purdue University |
| Jing Zhang | Purdue University |
| Julian Vuong | Purdue University |
| Shaiv Mehra | Purdue University |
| Craig J Goergen | Purdue University |
| David Cappelleri | Purdue University |
| Joseph Wallace | Purdue University |
| Kevin Otto | Purdue University |
| Meghaj Kabra | Purdue University |
| Katerina Grigoriou | Purdue University |
| Sooyeon Jeong | Purdue for Life Foundation |
| Darrell Pirtle | Purdue University |
| Tridib K. Saha | Purdue University |
| Julia Sibley | Purdue University |
| Maxwell Kawada | Purdue University |
| Ramses Martinez | Purdue University |
| Juan Wachs | Purdue University |
| Behzad Esmaeili | Purdue University |
| Stephan Biller | Purdue University |
| Denny Yu | Purdue University |
| Young-Jun Son | Purdue University |
| Yuehwern Yih | Purdue University |
| Violet Frye | Purdue University |
| Shengfeng Yang | Purdue University |
| Pelumni Oluwasanya | Cook Medical |
| Jason Corso | University of Michigan |
| Timothy M Kowalewski | University of Minnesota |
| Ivanna Yllahuaman | Purdue University |

## 7. Acknowledgements

This work was partially supported by the Center for AI and Robotic Excellence in medicine (CARE) at Purdue University and the Edwardson School of Industrial Engineering.

# 9. Appendices

## Appendix A: Workshop CARE 2025 Post-Event Survey

### Survey

Below we present the survey itself as it was shared with the participants

Thank you for attending the CARE 2025 Workshop. Your feedback is essential for shaping the vision and strategic plan for the proposed Center for Automation, Robotics, and Engineering in Surgery (CARE). Your responses will help us assess community needs, identify research gaps, and provide evidence of demand for the center as part of an NSF planning grant submission.
The survey should take approximately 5–7 minutes.

* Indicates required question

1. What is your primary role? *
    - ○ *Academic researcher*
    - ○ *Clinician / Surgeon*
    - ○ *Industry professional*
    - ○ *Government / Federal agency representative*
    - ○ *Student or trainee*
    - ○ *Other:*

2. How did you participate in the CARE 2025 Workshop? *
    - ○ *In person*
    - ○ *Virtually*
    - ○ *Other:*

3. Overall, how satisfied were you with the workshop?*

| *Very dissatisfied* | | | | *Very satisfied* |
|---|---|---|---|---|
| *1* | *2* | *3* | *4* | *5* |

4. How relevant were the workshop topics to your professional interests? *

| *Not relevant* | | | | *Extremely relevant* |
|---|---|---|---|---|
| *1* | *2* | *3* | *4* | *5* |

5. How well did the workshop address current challenges in surgical robotics, AI, and medicine? *

| *Not relevant* | | | | *Very relevant* |
|---|---|---|---|---|
| *1* | *2* | *3* | *4* | *5* |

6. To what extent do you believe there is a national need for a center like CARE? *

*No need* *Critical national need*

| | | | | |
|---|---|---|---|---|
| *1* | *2* | *3* | *4* | *5* |

7. Which of the following research pillars should be core priorities for the CARE center? *
    - *Autonomous surgical systems*
    - *Human–robot interaction in surgery*
    - *AI for surgical planning and decision support*
    - *Data science, simulation, and digital twins*
    - *Safety, validation, and regulatory science*
    - *Workforce training, education, and human capital*
    - *Clinical trials and translational pathways*
    - *Other:*

1. Did the workshop help you better understand the vision and mission of the proposed CARE center?*

| | | | | |
|---|---|---|---|---|
| *Yes, significantly* | *Yes, somewhat* | *Neutral* | *Not really* | *Not at all* |

2. How likely are you to engage with or collaborate on future CARE center activities (e.g., research,
proposals, training, industry partnerships)?*

*Not likely* *Very likely*

| | | | | | | | | |
|---|---|---|---|---|---|---|---|---|
| *1* | *2* | *3* | *4* | *5* | *6* | *7* | *8* | *9* |

10. What gaps in the field of surgical robotics and AI should the CARE center prioritize?
*[Free response]*

11. What types of resources or support would be most valuable for your research or organization?
    - *Access to clinical data*
    - *Access to simulation environments or testbeds*
    - *Industry partnerships and commercialization pathways*
    - *Regulatory and policy guidance*
    - *Interdisciplinary collaboration opportunities*
    - *Educational and workforce development programs*
    - *Funding or seed grants*
    - *Other:*

12. Did the workshop facilitate new collaborations or connections for you? *
    - *Yes*

- *Somewhat*
- *No*
- *Not sure*

13. Please describe any collaborations, ideas, or initiatives that emerged for you during the workshop.

*[Free response]*

14. What aspect of the workshop was most valuable for demonstrating the need for a national center like CARE?

*[Free response]*

## Responses

1. What is your primary role?
23 responses

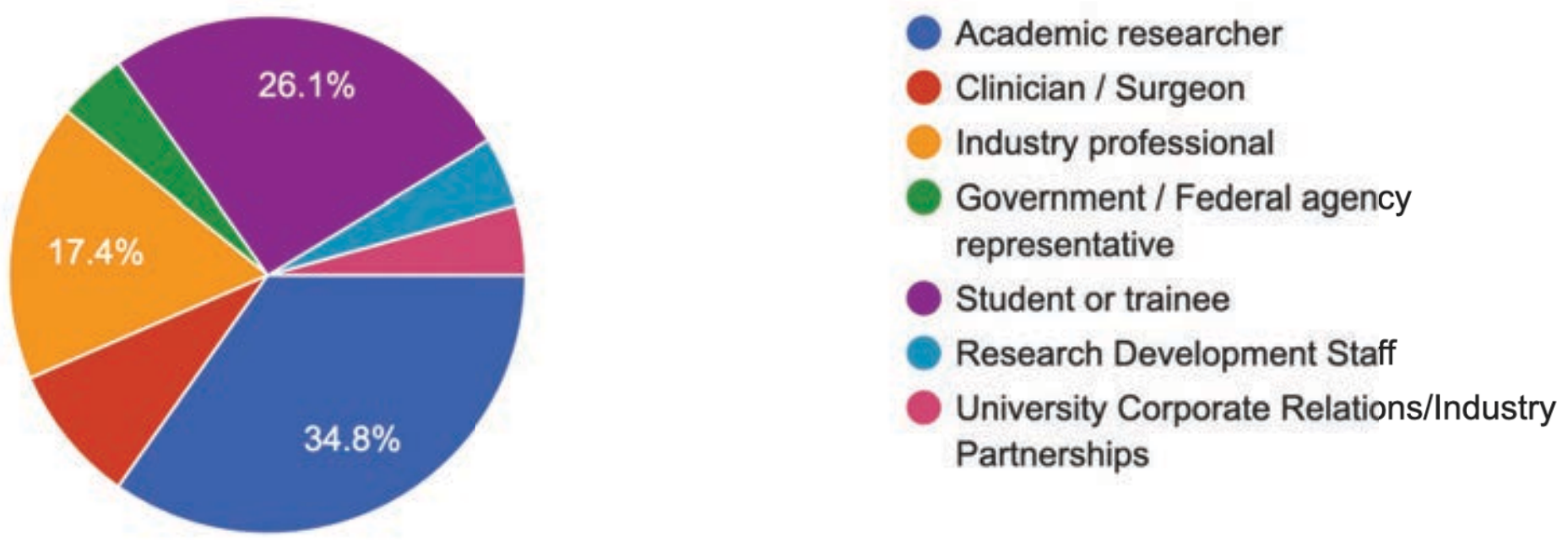

2. How did you participate in the CARE 2025 Workshop?
23 responses

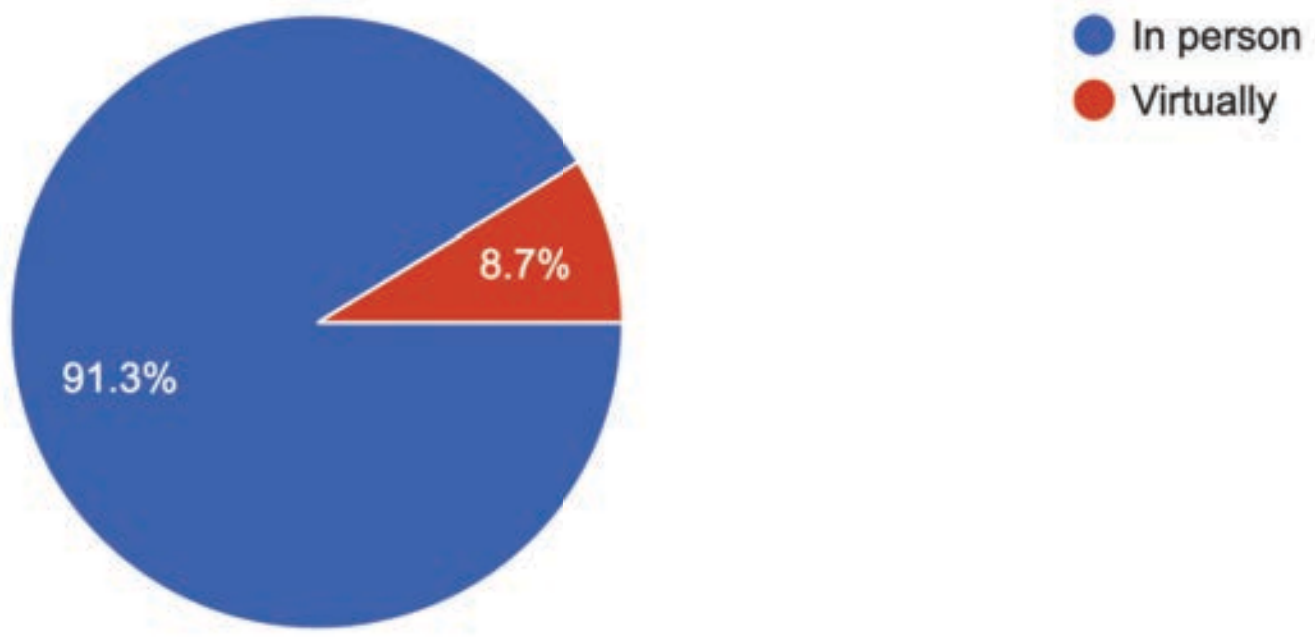

3. Overall, how satisfied were you with the workshop?
23 responses

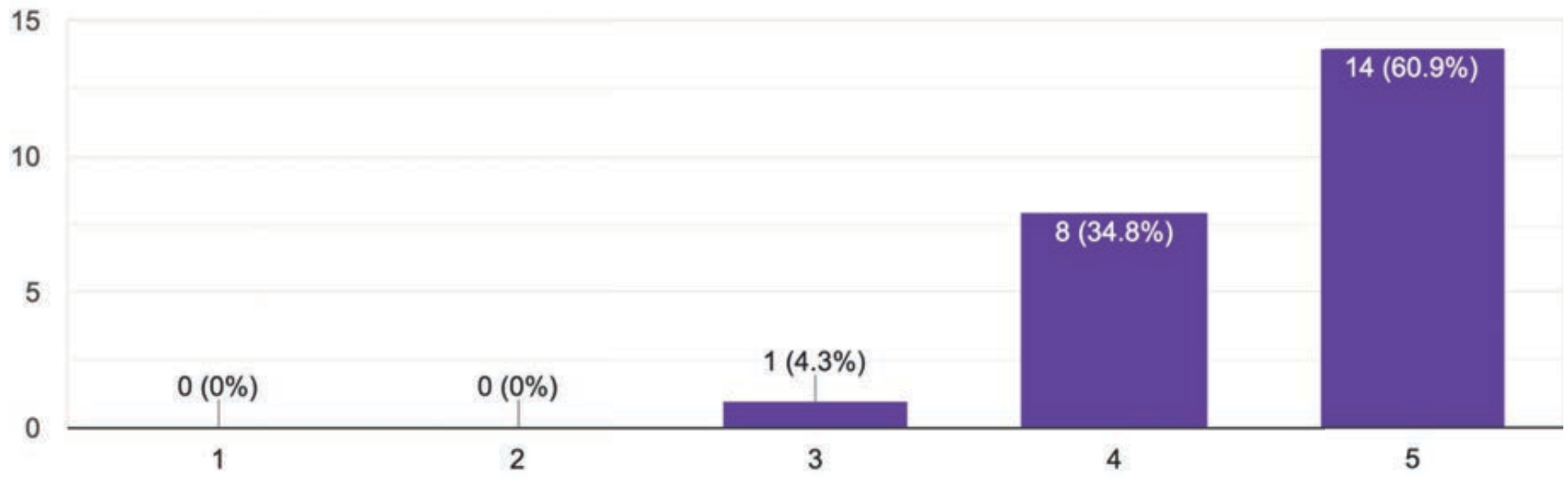

4. How relevant were the workshop topics to your professional interests?
23 responses

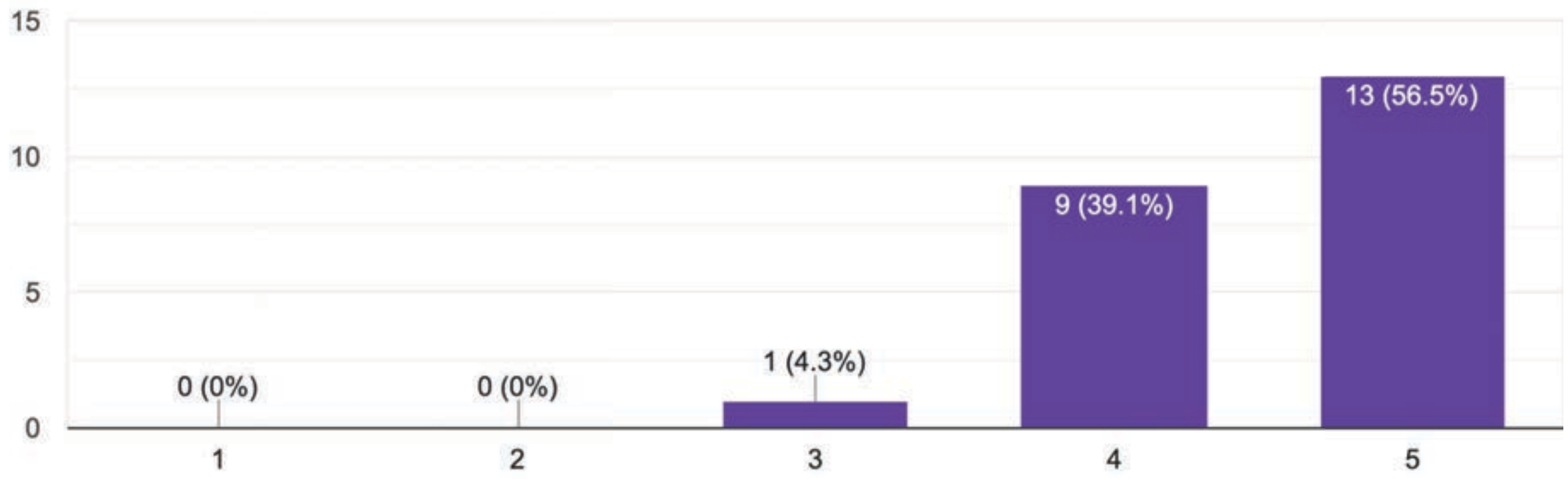

5. How well did the workshop address current challenges in surgical robotics, AI, and medicine?
23 responses

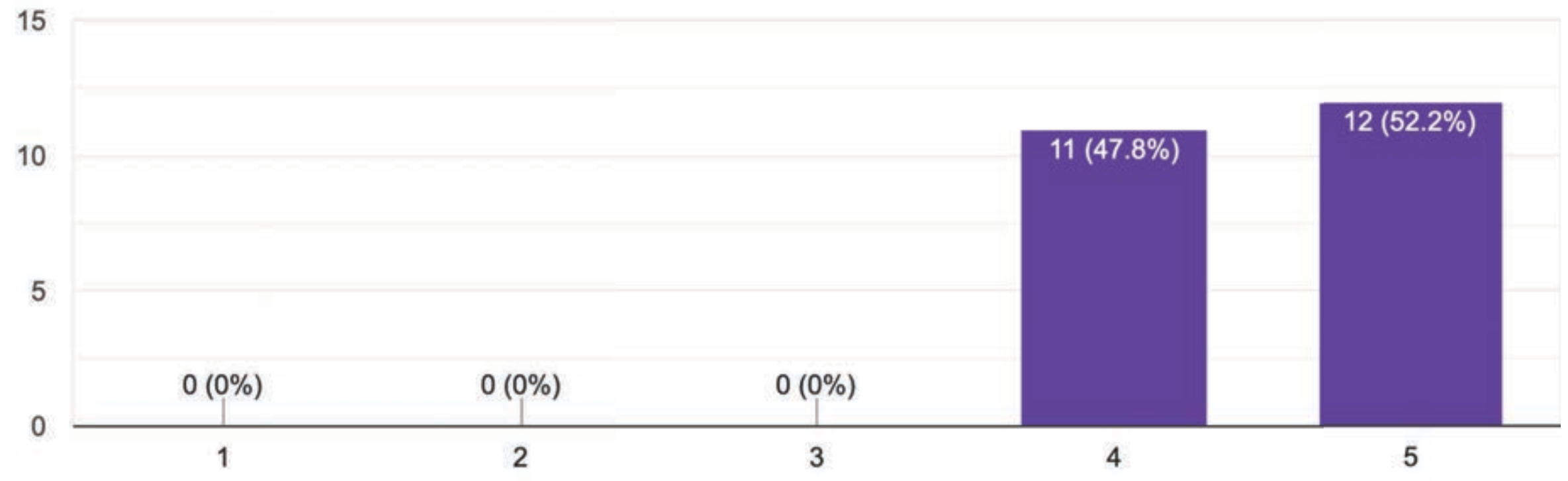

6. To what extent do you believe there is a national need for a center like CARE?

23 responses

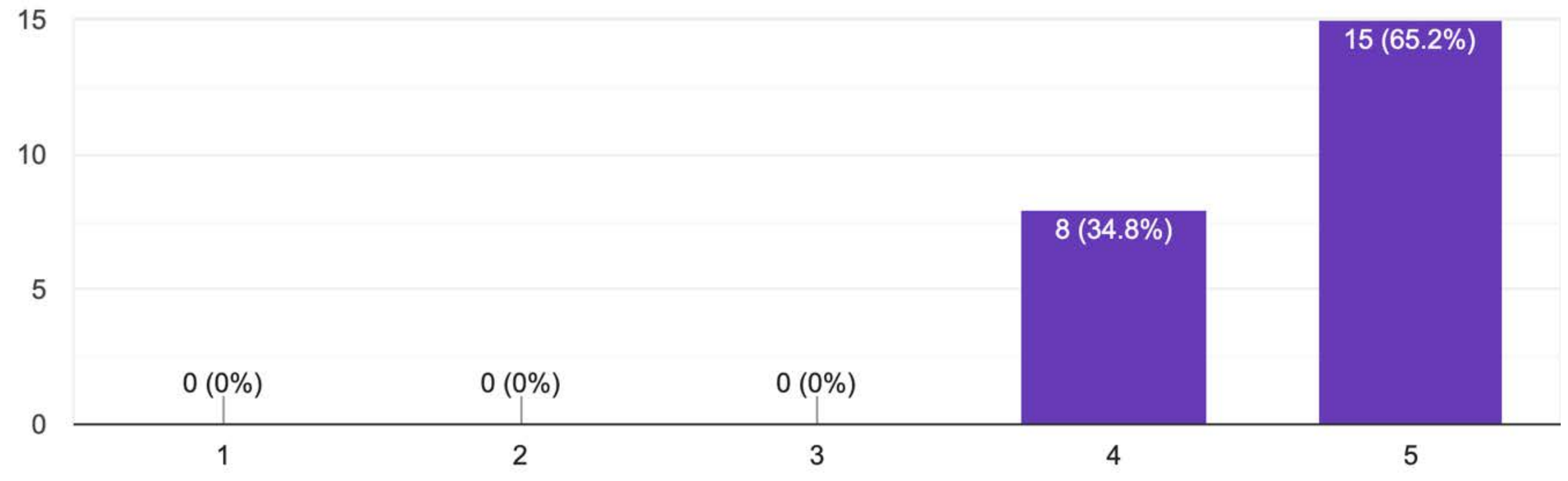

7. Which of the following research pillars should be core priorities for the CARE center?

23 responses

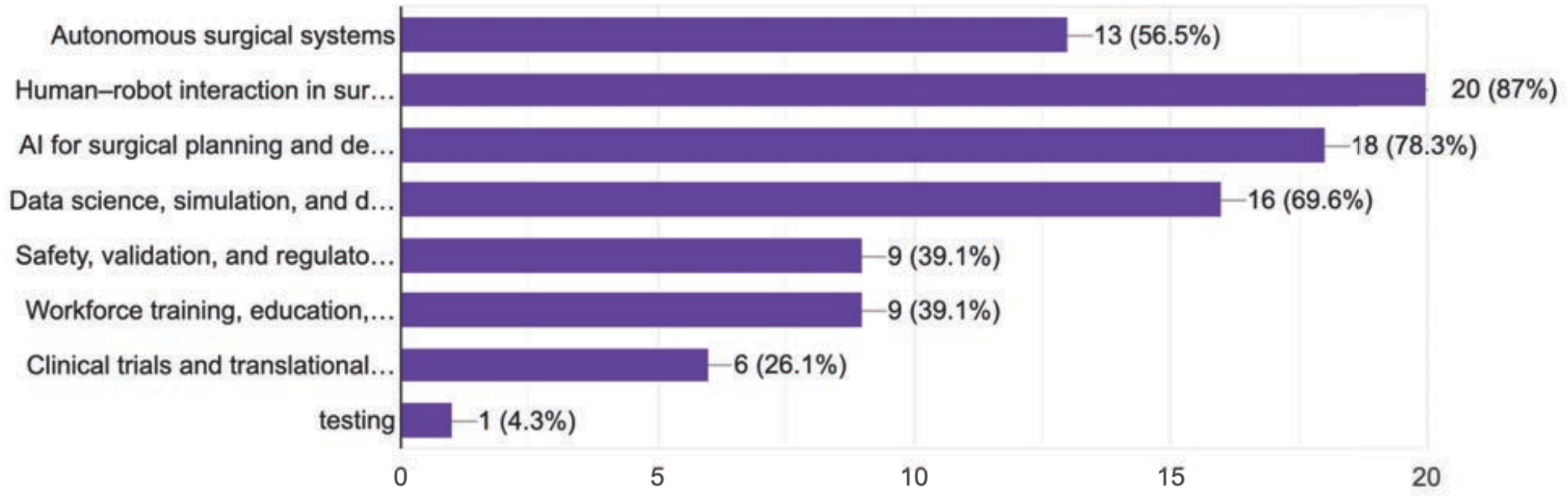

8. Did the workshop help you better understand the vision and mission of the proposed CARE center?
23 responses

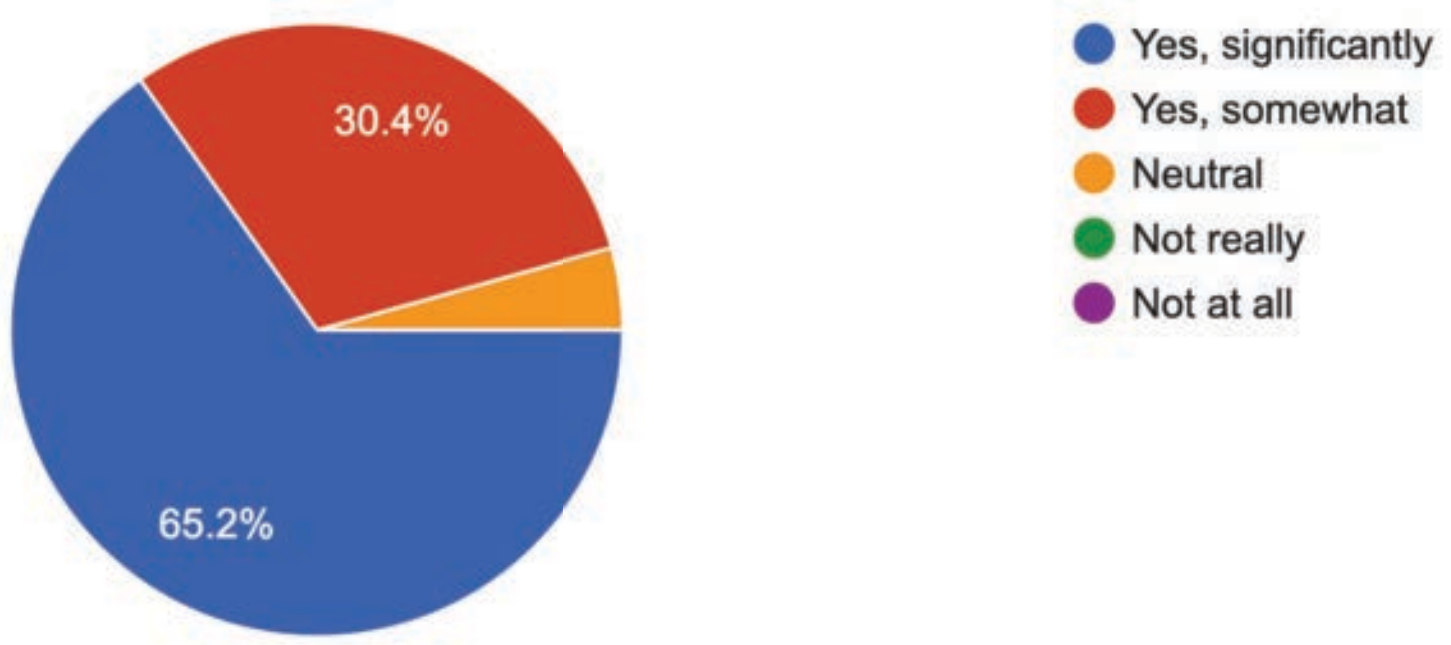

9. How likely are you to engage with or collaborate on future CARE center activities (e.g., research, proposals, training, industry partnerships)?
23 responses

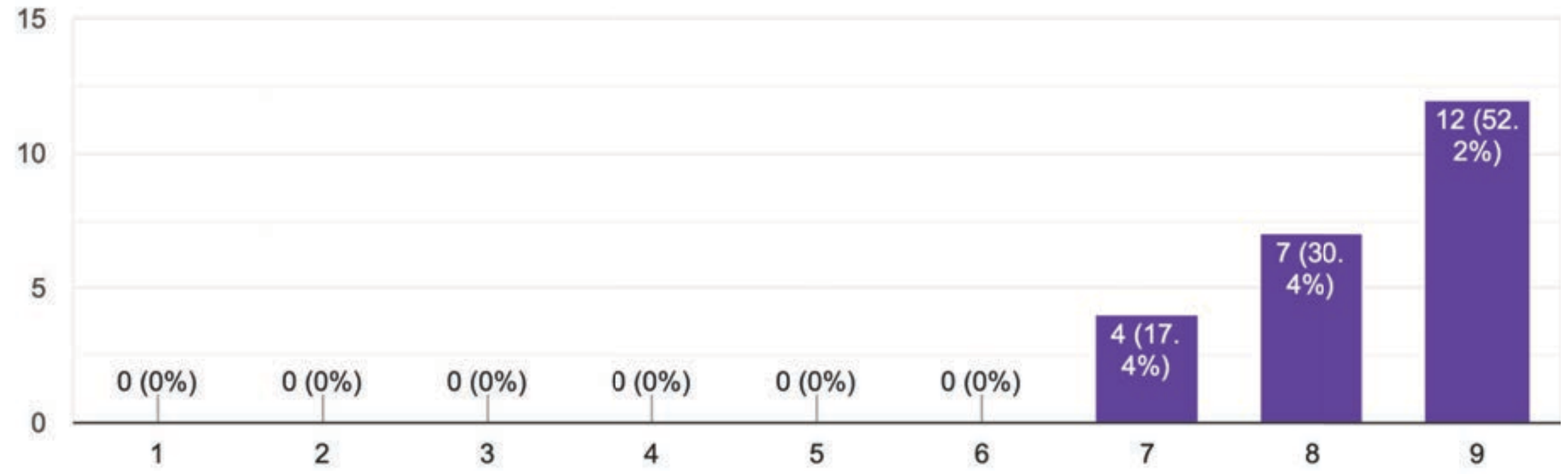

10 . What gaps in the field of surgical robotics and AI should the CARE center prioritize?
11 responses

- Safety, bridging simulation to reality, exploring possibilities in pain science/management;
    - Complete core curriculum for robotic surgery training, (using 'benchmarks' set by your surgical faculty) that the learners (residents, fellows and practicing surgeons) must meet in order to get privileges to perform robotic surgery on patients.
    - Once the CARE is established, apply to the American Collage of Surgeons - Accredited Educational Institutes (ACS-AEI) for certification privileges. Go to the ACS-AEI website NOW and there is an application section where there is

enormous amount information (especially about administration and curriculum development) about setting up a surgical simulation program.
    - Attend the Annual ACS-AEI Surgical Simulation Summit in Chicago each March.
- How to enhance existing surgical systems in surgical task semi-automations (surgical case/anatomy specific) using Multiphysics simulations, digital twins and Robot learning Policies. How can it be translated to surgical robotic setups (da Vinci framework) for pre-clinical testing on soft tissue Phantoms or cadavers for baseline measures and establishment of ground truth for AI assisted optimal surgical skills learning.
- CARE should prioritize accessibility for such technology to be used in academic research.
- Data integration approaches, Autonomous systems, adaptive robotics, robot human interactions
- More Human Factors
- human-AI/machine/robot teaming and collaboration
- AI assistant, autonomous instruction system, autonomous surgical guidance, autonomous surgery
- Robots and AI for Trauma Care
- Autonomous tasks for Soft tissue robotic surgery
- I would like more depth into surgical robotic assistants/nurses. While not surgery related, it would be very interesting to dive into AI and robotics in assistive technology (still within medicine).

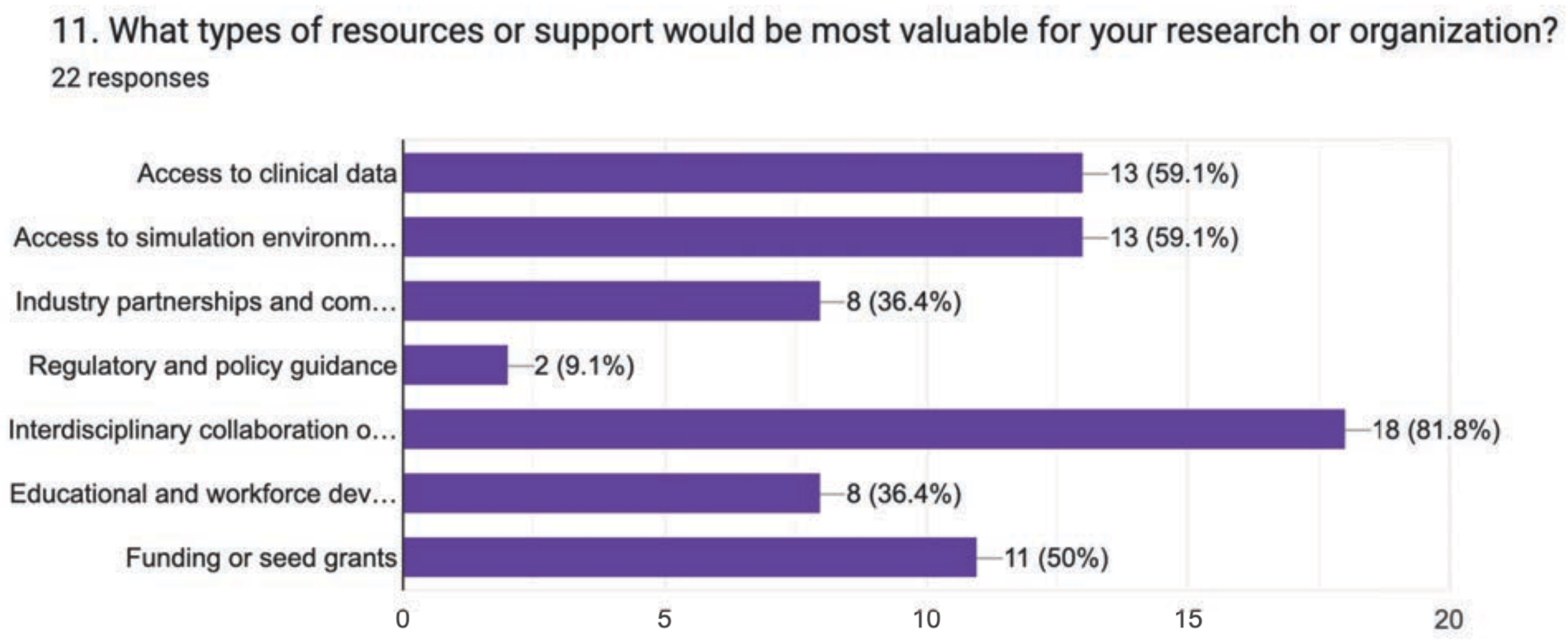

12. Did the workshop facilitate new collaborations or connections for you?
23 responses

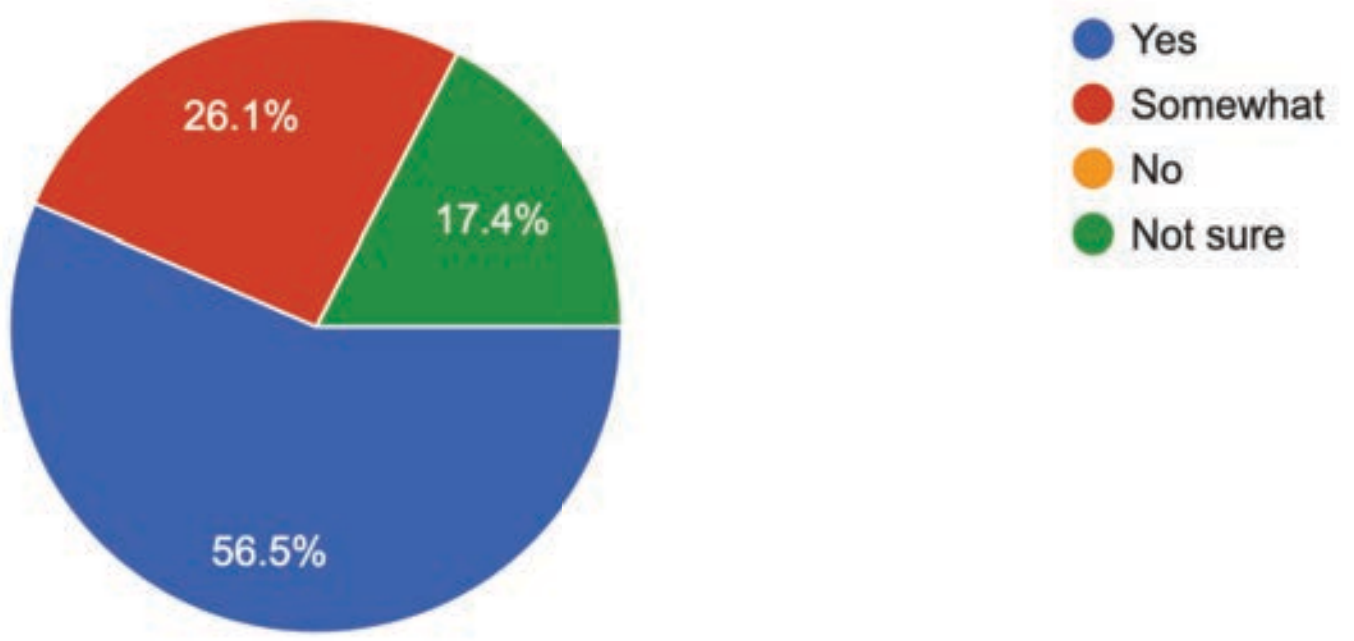

13. Please describe any collaborations, ideas, or initiatives that emerged for you during the workshop.
10 responses

- Making connections outside of my field (attended as an AI researcher in the aerospace industry and I believe there is a space for cross-learning between professionals from other industries).
- As above - first visit and then join ACS-AEI on the ACS website - much valuable information
- I received interests from Dr. Tim Kowalewski for further opportunities in LightSide Surgical, especially in Surgical Data simulation, advancing Multiphysics Engines and Digital Twins for enhancing Surgical AI and Robot motion planning.
- Confirm an AV technician next time... and demand more man power for the day of. :)
- It was informative to learn about other funding agencies priorities and how they differ from agency to agency.
- I missed most of the workshop because I could not attend due to flight cancellations. But I am interested to pursue them.
- conversation-based decision support for surgeons based on LLM system
- Learned more about DARPA/ARPA-H idea. Connected with other companies for potential collaboration. Presented our work to hopefully foster more collaboration.
- Networked with DVRK researchers, engineers, surgeons and gov't research employees. Plan to invite attendees to attend the Intuitive Surgical AI Innovators Summit in August.
- I have been able to connect with some of the professors and companies in attendance.

14. What aspect of the workshop was most valuable for demonstrating the need for a national center like CARE?

10 responses

- The inclusion of industry, academia, and government significantly conveyed the National need
- The presentation of the various attendees and their disciplines demonstrates the amount of work and collaboration that is needed
- The breakout rooms, the discussions and the summarization of all the possible necessities where AI can make improvements in existing Robot assisted Surgical technologies.
- Academia-Industry-Government interactions and collaborations.
- opportunities for improving rural health
- NA
- It was great to hear the core stakeholders' perspective on how current robotic systems could be improved to better support and standardize surgical procedures and quality for better patient outcomes
- All funding agencies mentioned their interest in funding AI assistants for real world medical care. There is a big issue of trust for this kind of critical scenarios. Currently, they don't want AI to make decisions, but to provide guidances or supports. There's a huge gap in this area.
- The workshop was attended by a diverse group of experts and learners. The CARE Center will provide an essential collaborative environment for research partnerships.
- The discussion groups were really powerful as it allowed for the attendees to interact ad brainstorm together.

## Post Survey Analysis & Executive Summary

*CARE 2025 Workshop Survey Results*
Source: *Responses to the Survey*
Total Respondents: 23

The CARE 2025 Workshop survey results demonstrate overwhelming support for establishing a national center focused on surgical robotics and artificial intelligence (AI). Participants expressed high satisfaction with the workshop, strong belief in the national need for a CARE center, and significant interest in future collaboration. The findings also identify clear technical priorities, infrastructure gaps, and strategic opportunities.

*Participant Profile*

Respondents represented a diverse cross-section of stakeholders:

- **Academic researchers (34.8%)** – largest group
- **Students/trainees (26.1%)**
- **Industry professionals (17.4%)**
- Additional representation from clinicians, government, and research staff
- **91.3% attended in person**, indicating strong engagement

This diversity reflects broad cross-sector interest in advancing surgical robotics and AI.

*Workshop Impact & Reception*

Feedback was overwhelmingly positive:

- **95.7% rated overall satisfaction as 4 or 5 (out of 5)**
- **95.6% rated workshop relevance as 4 or 5**
- **100% agreed (rated 4 or 5) that there is a national need for a CARE center**
- **100% rated the workshop's coverage of current challenges as 4 or 5**
- Over **82% reported a high likelihood (8–9/9) of future engagement**

These results signal strong momentum and readiness for continued development of CARE initiatives.

*Priority Research Areas*

Participants identified the following as core priorities for the CARE center:

1. **Human–robot interaction in surgery (87%)**
2. **AI for surgical planning and decision-making (78.3%)**
3. **Data science, simulation, and modeling (69.6%)**
4. **Autonomous surgical systems (56.5%)**
5. Safety, validation, and regulation (39.1%)
6. Workforce training and education (39.1%)

There is clear emphasis on advancing AI-enabled autonomy while maintaining strong human-AI collaboration.

*Key Gaps Identified*

Open-ended responses highlighted several strategic gaps:

<u>1. Simulation-to-Clinical Translation</u>

- Bridging simulation and real-world deployment
- Digital twins and multi-physics modeling
- Benchmarking and ground-truth validation

2. AI Assistance & Autonomy

- AI surgical assistants and semi-autonomous tasks
- Conversational decision support systems (LLM-based)
- Trauma-care robotics and autonomous guidance

3. Human Factors & Trust

- Human-AI teaming frameworks
- Addressing trust and decision authority in clinical settings

4. Training & Credentialing

- Standardized robotic surgery curricula
- Benchmark-based certification pathways
- Alignment with national accreditation bodies

### Resource Needs

The most requested forms of support were:

1. ***Interdisciplinary collaboration infrastructure (81.8%)***
2. Access to clinical data (59.1%)
3. Access to simulation environments (59.1%)
4. Funding/seed grants (50%)

Respondents clearly value CARE as a collaborative hub enabling data access, translational research, and cross-sector coordination.

### Strategic Implications

The survey results confirm:

- Strong national demand for a centralized center advancing surgical robotics and AI
- High stakeholder readiness to collaborate
- Clear priorities around autonomy, simulation, AI-human teaming, and validation
- A need for shared infrastructure, training standards, and translational pathways

The CARE center is well-positioned to serve as a national convener—integrating academia, industry, and government to accelerate safe, trustworthy, and clinically impactful AI-enabled surgical technologies.

## Appendix B - Workshop Agenda

| | | |
|---|---|---|
| 8:30 - 9:00 | **Arrival & Networking (Light breakfast)** | |
| 9:00 - 9:30 | **Welcome & Introductions** | |
| | Juan Wachs | |
| | Dr. Karl Billimoria | |
| | Joey Wallace | |
| 9:30 - 10:15 | **Keynote 1** | |
| | Rick Satava (U of W) | |
| 10:15 - 10:30 | **Break** | |
| 10:30 - 11:30 | **Industry Lightning Presentations - 10 min talk, 5 Q&A** | |
| | Gordon Wisbach | Intuitive Surgical |
| | Tony Romano | Zimmer Biomet, Knees |
| | Tim Kowalewski | Lightside Surgical |
| | Rachel Clipp | Kitware |
| 11:30-12:30 | **Federal Panel , 5-10 min talk tops, Q& A from Audience** | |
| | ***Moderator : Juan or Julia*** | |
| | Jeremy Pamplin | DARPA |
| | Shivani Sharma | NSF |
| | Tyler Best | ARPA-H |
| 12:30 - 1:30 | **Lunch** | |
| 1:30 - 2:15 | **Keynote 2:** | |
| | Jason Corso (U Michigan) | |
| 2:15 - 3:15 | **Academic Lightning Talks 10 min talk, 5 Q&A** | |
| | Denny Yu | Purdue |
| | Dimitrios Stefanidis | IUSM |
| | Dave Cappelleri | Purdue |
| | Andrew Gonzalez | IUSM - Clinical |
| 3:15 - 4:15 | **Working Group Discussions- 4 Breakout Rooms** | |
| 3:15 – 3:25 | Thought Leadership Talk, 5-10 minutes tops. | |
| 3:30 – 4:15 | BREAK items available | |
| 4:15 - 4:45 | ***Group Summary*** | |
| 4:45 - 5:00 | ***Closing Remarks*** | |
| 5:00 | **Networking Reception** | |

## Appendix C - Speakers' Abstracts and Biographies

### C.1 Richard Satava, MD FACS, PhD(hc)

Professor Emeritus of Surgery, University of Washington Medical Center

ABSTRACT:
Even as this fourth revolution in surgery in 25 years (robotic surgery) continues to gain in acceptance, a much more disruptive change is beginning as the next revolution, Artificial Intelligence, which is just the tip of the iceberg that heralds the transition to remote telesurgery for remote (transcontinental) surgery . When combined with other information systems technologies, imaging systems, Virtual Reality (VR), molecular and genetic manipulation, and nanotechnology (to name a few), diseases will also begin to be cured at the cellular and molecular level, and non-invasively. Such systems are based upon the premise that robotics, AI and automation can bring precision, speed and reliability, especially as surgery 'descends' into operating at the cellular and molecular level.
In addition when robotics combines with Artificial Intelligence (AI), the 5th and 6th generation (5G and 6G) telecommuications, supercomputing, and telesurgery, there will be an exponential increase in opportunities for innovation on a global scale. However, with these opportunities, there will also be significant challenges, not only technological, but also behavioral, humanitarian, political and ethical issues. The time has come to rethink what the future of robotics with AI can bring to surgery.

BIOGRAPHY:
Prior academic positions include Professor of Surgery at Yale University and a military appointment as Professor of Surgery (USUHS) in the Army Medical Corps assigned to General Surgery at Walter Reed Army Medical Center and former Astronaut candidate. Government positions included Program Manager of Advanced Biomedical Technology at the Defense Advanced Research Projects Agency (DARPA) for 12 years and Senior Science Advisor at the US Army Medical Research and Materiel Command in Ft. Detrick, Maryland, and Director of the NASA Commercial Space Center for Medical Informatics Telemedicine, and Advanced Technology (NASA-CSCMITAT) at Yale University. Upon completion of military career and government service he had continued clinical medicine at Yale University and University of Washington. He also holds a PhD(hon) at Semmelweis University in Budapest, Hungary and PhD(hon) at Titu Maiorescu University in Bucharest Romania and DSc from Medical University of Pleven, Pleven Bulgaria.

He has served in government on the White House Office of Science and Technology Policy (OSTP) Committee on Health, Food and Safety and was also awarded the prestigious Department of Defense Legion of Merit and Department of Defense Exceptional Service medals as well as awarded the Smithsonian Laureate in Healthcare. He has been a member of numerous committees of the American College of Surgeons (ACS), currently serving on the ACS-Accredited Education Institutes (ACS-AEI). He is a Past President of the Society of American Gastrointestinal Endoscopic Surgeons (SAGES), the Society of Laparoendoscopic Surgeons (SLS), the Society of Medical Innovation and Therapy (SMIT), and a former member of the Aerospace Medical Association. He was a member of the National Board of Medical Examiners (NBME) and is currently on the Board of many surgical societies and on the editorial board of numerous surgical and scientific journals, and active in a number of surgical and engineering societies.
In pioneering research in telepresence surgery, he was the surgeon on the project that developed the first surgical robot, which later became the DaVinci Surgical Robot. He also was the founder of the Medicine Meets Virtual Reality (MMVR) conference and built (with Jaron Lanier), the first VR simulator for surgery (in 1989). Shortly thereafter, while at DARPA, he funded all robotic surgery research and all VR medical simulation for the first 10 years of their development.

For 5 years he was a member of the Advisory Board of the National Space Biomedical Research Institute (NSBRI) advising NASA in the use of advanced biometric sensing, haptics and other life science research for astronauts. Now Dr. Satava has added being continuously active in surgical education and surgical research, with more than 250 publications and book chapters in diverse areas of advanced surgical technology, including, Video and 3-D imaging, Plasma Medicine, Directed Energy Surgery, Telepresence Surgery, Robotic Surgery and telesurgery, Virtual Reality Surgical Simulation, Objective Assessment of Surgical Competence and Training, Surgical applications in AI, Surgery in Space, and the Moral and Ethical Impact of Advanced Technologies.

During his 23 years of military surgery he had been an active flight surgeon, an Army astronaut candidate, combat tours of duty as MASH surgeon for the Grenada Invasion, and a hospital commander during Desert Storm, all the while continuing clinical surgical practice. Current research is focused on advanced technologies to formulate the architecture for the next generation of clinical Medicine and Surgery, education and training, telesurgery and Surgical AI and Surgery in Space.

## C.2 Gordon Wisbach, MD, MBA, Capt, MC, USN (Ret)

Intuitive Surgical

ABSTRACT:
Advances in artificial intelligence (AI) are reshaping the future of minimally invasive surgery, offering new opportunities to enhance precision, safety, and clinical decision-making. This presentation highlights Intuitive Surgical’s latest innovations integrating AI-driven capabilities across the surgical continuum—from preoperative planning and intraoperative guidance to postoperative analytics. We will review current applications, including automated tissue identification, real-time procedural insights, and workflow optimization within the da Vinci ecosystem. Emerging research in machine learning, computer vision, and data-enabled training will also be discussed. Together, these developments illustrate how AI can support surgeons, elevate performance, and improve patient outcomes in the next generation of robotic surgery.

BIOGRAPHY:
Gordon Wisbach, MD, MBA is a general surgeon that specializes in Minimally Invasive, Metabolic/Bariatric as well as Robotic surgery at the Navy Medicine Readiness & Training Command San Diego (NMTRC-SD). He founded the ACS-accredited Surgical Simulation/Education Fellowship and was the inaugural Tele-Surgical Director of the Virtual Medical Operations Center. Dr. Wisbach was awarded his Medical Degree from Jefferson Medical College in Philadelphia, Pennsylvania and completed his residency training at NMRTC-SD. He is fellowship trained in Advanced Laparoscopic/Bariatric Surgery at Brigham & Women's Hospital in Boston, Massachusetts and earned his MBA from the Naval Post-graduate School in Monterey, California. He is a Professor of Surgery at the Uniformed Services University of the Health Sciences in Bethesda, Maryland and, in 2022, CAPT Wisbach honorably retired from the US Navy after 24 years of service. During his accomplished career, he served successful leadership positions including inaugural military-wide Director of Surgical Services and carried out numerous overseas deployments that ranged from leading humanitarian to combat surgery teams in austere environments. After military retirement, Dr. Wisbach joined Intuitive Surgical full-time as Lead for Clinical Integration and Telecollaboration to develop digital surgery under the premise to design products and services "by surgeons for surgeons". Dr. Wisbach has active research interests in surgical education using simulation and advancing surgical tele-mentoring on a trajectory towards tele-surgery as well as leveraging surgical data science to optimize the delivery of surgical care.

## C.3 Tony Romano

Associate Director, Zimmer Biomet

ABSTRACT

Advances in artificial intelligence and surgical robotics are transforming the future of total knee arthroplasty (TKA), offering new opportunities to understand and optimize the key drivers of patient outcomes and satisfaction. This presentation explores how AI-enabled robotic platforms can move beyond traditional mechanical alignment goals to provide a data-driven, patient-specific approach to knee reconstruction.

As robotic systems enhance the precision of implant placement and enable real-time assessment of knee position, soft-tissue behavior, and intraoperative laxity, they generate an unprecedented depth of quantitative information. When combined with AI, these data streams allow surgeons and product teams to examine which variables—alignment, balance, patient morphology, or functional kinematics—most strongly influence postoperative recovery and long-term satisfaction.

The session will address the ongoing debate between alignment-focused and balance-focused philosophies, highlighting how AI can help reconcile these approaches by identifying optimal combinations tailored to individual patients. By integrating robotics-driven accuracy with AI-assisted decision support, the next generation of TKA technologies has the potential to improve consistency, personalize surgical strategy, and ultimately elevate the standard of orthopedic care.

BIOGRAPHY

Tony Romano is an experienced product management and engineering leader with over 20 years at Zimmer Biomet, specializing in robotic and knee arthroplasty technologies. He currently serves as Associate Director of Product Management, overseeing strategy, development, and commercialization for robotic total knee solutions.

His prior roles include Associate Director of Supplier Excellence, Integration Manager, Product Manager for Patient Specific and Robotic Knees, Project Manager for New Products, and Senior Development Engineer. Across these positions, he has contributed to major knee system programs including Persona, Natural Knee Flex, and Patient Specific Instruments.

Tony holds a B.S. in Mechanical Engineering from Purdue University and an MBA from Indiana Institute of Technology, supporting his ability to integrate technical innovation with business strategy. He is also the author of the chapter "Unicompartmental Arthroplasty with Patient Specific Instruments" in Improving Accuracy in Knee Arthroplasty (Jaypee Brothers Medical Publishing, 2012).

## C.4 Tim Kowalewski, PhD

Lightside Surgical
Associate Professor, *University of Minnesota*
Chief Technical Officer, Co-Founder, *LightSide Surgical*

## C.5 Rachel Clipp, PhD

Assistant Director of Medical Computing
*Kitware, Inc.*

ABSTRACT
*"Towards a Human Physiological Digital Twin with the Pulse Physiology Engine"*

Kitware is an open-source research and development company with a focus on artificial intelligence. We work on projects in the medical space across a range of technical areas including computational physiology, medical triage, image analysis and segmentation, surgical simulation, and cyber-physical systems. Dr. Clipp will provide an overview of Kitware's expertise in these areas and highlight specific projects related to these areas.

BIOGRAPHY

Rachel Clipp, Ph.D., is a medical computing expert on Kitware's Medical Computing Team located in Carrboro, North Carolina. She conducts research in computational modeling and artificial intelligence as applied to biomedical problems. Rachel leads medical modeling and simulation projects at Kitware, including those that involve the open-source Pulse Physiology Engine. She also leads computational modeling projects using Lattice Boltzmann Methods and artificial intelligence models that represent outcomes from high-fidelity physics-based models and to predict life-saving interventions. She is funded by projects from the NIH, DARPA, DoD, and industry partners.

Under Rachel's guidance, Pulse has been successfully incorporated in commercial and government-funded products and programs. The Pulse team has addressed the needs of the military for virtual medical simulation through collaborations with Exonicus to develop the Trauma Simulator and SimQuest and BioMojo to contribute to the Modeling and Simulation Training Architecture. They have also collaborated with academic and clinical institutions to test medical device algorithms with a closed-loop physiology management system. Kitware funded a project using Pulse to study the use of ventilators for multi-patient treatment in the early stages of the COVID-19 pandemic. Current projects include hemorrhage and fluid resuscitation modeling, ECMO modeling, synthetic data generation, and developing triage recommendations using large language models.

Rachel's graduate work focused on the development of dynamic boundary conditions for use in finite element analysis and computational fluid dynamics. The boundary conditions developed were used to predict the effects of respiration on the pulmonary vasculature. She also developed a benchtop apparatus to perfuse and ventilate excised lamb lungs to collect hemodynamic and respiratory data for validation of the dynamic boundary conditions.

Rachel received her Ph.D. and master's degree in biomedical engineering from the University of North Carolina at Chapel Hill and North Carolina State University. She received her bachelor's degree in mechanical engineering from Clemson University.

## C.6 Col. Jeremy Pamplin, MD

Program Manager, Biological Technologies Office,
*DARPA*

## C.7 Shivani Sharma, PhD

**Program Director,**
***NSF***

BIOGRAPHY
Dr. Shivani Sharma is a Program Director in National Science Foundation's Division of Civil, Mechanical, and Manufacturing Innovation under the Directorate for Engineering. She oversees a portfolio of Biomechanics & Mechanobiology program, NSF-funded research Centers including the Science & Technology Center for Engineering Mechano-Biology, and supports inter-agency programs such as the Smart Health and Biomedical Research in the Era of Artificial Intelligence and Advanced Data Science (SCH). Dr. Sharma champions the integration of artificial intelligence, computational modeling, and robotics into mechanobiology and biomedical research through strategic national and international partnerships.

## C.8 Tyler Best

Acting Director, Health Science Futures Office,
*Advanced Research Projects Agency for Health (ARPA-H)*

ABSTRACT
What if we could perform surgery autonomously? Surgery today is tethered to skilled practitioners with specialized training using manual approaches. Endovascular interventionists, for example, must perform intricate maneuvers of rudimentary devices through complex, delicate vessels using only static models and occasional 2D imaging. The availability of experts becomes paramount, creating impediments to care that are acutely felt in remote regions by patients of urgent conditions such as stroke. Contemporary technologies lack the perception, recognition, and therapeutic capabilities to treat patients autonomously, away from the active attention of a specialist.
The Autonomous Interventions and Robotics (AIR) program launched by ARPA-H aims to catalyze the development of autonomous surgical robots. Systems that can perform parts, or all, of a procedure without direct human input can have a significant impact in the way surgery is provided, performed, and perceived. AIR approaches critical gaps in the availability and reach of surgery from two distinct angles. Technical Area 1 focuses on endovascular robotics to treat urgent conditions such as stroke where patient outcomes are affected by the availability of specialists. Technical Area 2 focuses on microbots—miniaturized and untethered devices—to enable new paradigms of surgery such as those performed in clinics under minimal supervision. Leveraging bold approaches and cross-disciplinary expertise, ARPA-H hopes to mobilize the shared ingenuity of the community to transform surgical care.

BIOGRAPHY
Dr. Tyler Best joined ARPA-H in October 2022 from the National Institutes of Health, where he was a Program Officer in the Office of the Director. He is a distinguished neurotechnologist specializing in the development and management of high-reward scientific initiatives. With a rich background in both scientific research and program management, Best excels at identifying, clarifying, and implementing innovative solutions across government, industry, and academic sectors. His expertise spans peripheral nerve interfaces, neuronal biophysics, and various facets of molecular and cellular biology. Best's career includes significant contributions as a Booz Allen Hamilton associate, where he supported the Defense Advanced Research Projects Agency (DARPA). He earned a PhD in neuroscience from the Uniformed Services University and conducted postdoctoral work at Northwestern University and Urogenix Inc.

## C.9 Jason Corso, PhD

**Toyota Professor of AI,**
***University of Michigan***

ABSTRACT
"Hey Siri, Can you measure my left ventricle ejection fraction?" Despite having full control over the zeitgeist, Artificial Intelligence (AI) has yet to live up to its promise in many fields. Medicine, in particular, has significant upside potential with AI—with hospitals closing, a shortage of physicians and other medical professionals, and an instrinsically difficult domain, AI has a unique opportunity to upskill practice, bringing better care and better training to everyone. This talk will explore the problem, potential value, and early methods in upskilling medical practitioners along two axes. First, I will describe how visual AI methods are already impacting the cardiothoracic surgical domain via technical and non-technical assessment for more objective training and review. Second, I will describe how interactive, physically-grounded AI guidance can upskill medical practitioners and bring state of the art care into rural settings, which are among the most challenging settings for healthcare delivery. Ultimately, this talk will provide concrete evidence of the potential AI has in upskilling medical practice.

BIOGRAPHY
Dr. Jason J. Corso is a Professor of Electrical Engineering and Computer Science at the University of Michigan. He received his Ph.D. in Computer Science from Johns Hopkins University in 2005. His research spans computer vision, robotics, and machine learning. He is a recipient of the NSF CAREER Award, ARO Young Investigator Award, and Google Faculty Research Award, and serves on the DARPA CSSG.

He is also the Co-Founder and CEO of Voxel51, a computer vision startup building state-of-the-art platforms for video and image-based applications.

## C.10 Denny Yu, PhD

Associate Professor, *Purdue University*
Adjunct Associate Professor Surgery, *IU School of Medicine*

ABSTRACT
Biobehavioral sensing, e.g., computer vision and physiological monitoring, offers real-time, objective understanding into human cognitive states and behaviors. Integrating this information into robotic and AI systems enables human-aware technologies that dynamically respond to users' physical state, cognitive load, and situation awareness to support decision-making. This talk will explore opportunities and challenges in biobehavioral AI across healthcare applications.

BIOGRAPHY
Denny Yu, PhD, CPE is an Associate Professor of Industrial Engineering at Purdue University and an Adjunct Associate Professor of Surgery at Indiana University School of Medicine. He was previously a Fulbright Scholar (2023-2024) and a 2017-2023 Summer Faculty Fellow at the Air Force Research Laboratory's 711th Human Performance Wing. He also serves on the board of directors for the Board of Certification in Professional Ergonomics, Institute of Industrial and Systems Engineers Work Systems Division, and the Foundation for Professional Ergonomics.

Dr. Denny Yu's research focus on human factors engineering in the area of biobehavioral sensing techniques for human state modeling in high-stress, dynamic healthcare environments. His contributions include: 1) Objective and automated biobehavioral models for assessing physical capability, cognitive states, and team non-technical skills and 2) Real-time, user cognitive state-aware systems for enhancing human performance in high-workload environments. In conducting this work, his team has received the RSJ/KROS Distinguished Interdisciplinary Research Award (RO-MAN 2021), the 2021 Human Factors Prize (by the Human Factors and Ergonomics Society), the Young Investigator Award from the Applied Ergonomics Society, 2023 IEA/Tsinghua Award for Collaborative Human Factors and Ergonomics Education, the 2024 International Ergonomics Association's (IEA) Outstanding Educator Award, and the Fulbright Scholar award.

## C.11 Dimitrios Stefanidis, MD, PhD

Harris B. Shumacker Jr. M.D. Professor of Surgery
Director, MIS/Bariatric Surgery
Director, Department of Surgery Skills Lab
Indiana University

ABSTRACT
*"Artificial Intelligence in Surgical Education: Transforming Training, Assessment, and Skill Development"*

Surgical education faces several challenges related to the introduction of new technologies and techniques, increasing complexity of surgical patients, scrutiny of surgical performance, and optimization of patient outcomes. Artificial intelligence promises to overcome many of these challenges by introducing solutions not available to surgical educators previously. Advances in large language models, computer vision, and deep learning, offer unique opportunities to augment surgical education by enabling objective skill assessment, timely feedback provision on trainee performance, intraoperative guidance, training need identification, decision aid support, simulation training enhancement, and resource utilization. In this presentation, we will review promising applications of AI in surgical education and identify the most pressing needs for AI tool development to enhance surgical training.

### C.12 David Cappelleri, PhD

Assistant Vice President for Research Innovation
B.F.S. Schaefer Scholar & Professor
*Purdue University*

ABSTRACT
In this talk, I will highlight some recent advances in microrobotic surgical applications, providing examples of different types of families of wireless mobile microrobots driven by external fields for biomedical applications. I will then discuss future opportunities where AI can enhance microrobotic surgical applications in nearly every stage – from design and control to real-time operation and post-procedure analysis.

BIOGRAPHY
David J. Cappelleri is the Assistant Vice President for Research Innovation in the Office of Research, B.F.S. Schaefer Scholar & Professor in the School of Mechanical Engineering, and Professor in the Weldon School of Biomedical Engineering (by courtesy) at Purdue University. Professor Cappelleri founded the Multi-Scale Robotics & Automation Lab (MSRAL) that performs cutting-edge research on robotic and automation systems at various length scales. His research interests include mobile microrobotics for biomedical and manufacturing applications, surgical robotics, automated manipulation and assembly, and agricultural robotics. Prof. Cappelleri is the Purdue site director for the NSF Engineering Research Center on the Internet of Things for Precision Agriculture (IoT4Ag).

Prof. Cappelleri has received various awards, such as the NSF CAREER Award, Harvey N. Davis Distinguished Assistant Professor Teaching Award, the Association for Lab Automation Young Scientist Award, and is Fellow of the American Society of Mechanical Engineers (ASME). He received a Batchelor of Mechanical Engineering degree from Villanova University, a MS in Mechanical Engineering degree from The Pennsylvania State University, and a PhD degree in Mechanical Engineering & Applied Mechanics from the University of Pennsylvania.

## C13 Andrew Gonzalez, MD, JC, MPH

Assistant Professor of Surgery
Associate Director for Data Science and Research Scientist, Center for Health Services Research, Regenstrief Institute
*Indiana University*

ABSTRACT
Mimicking human clinician reasoning requires not only integrating multimodal data but also balancing efficiently training models to learn generalizable representations against clinical trust and verifiability. We discuss best practices for labelling, ontology development, and task formulation in the setting of interdisciplinary teams. We will explore opportunities to scale multimodal multitask learning within the CTSI ecosystem.

BIOGRAPHY
Andrew A. Gonzalez, MD JD MPH is an Assistant Professor of Vascular Surgery at the Indiana University School of Medicine and Associate Director for Data Science at the Center for Health Services Research at the Regenstrief Institute. Currently, his focus is using AI/ML to provide “the right information, at the right time, in the right format” to assist front line providers in making challenging decisions. Dr. Gonzalez served as the National Academy of Medicine Omenn Fellow (2021-23) and as a Diagnostic Excellence Scholar (2024-25). He is currently funded by the National Heart Lung and Blood Institute to develop “An intelligent clinical decision support system for peripheral arterial disease” (K23-HL-181388) and by the NIH CLINAQ initiative for “Adaptive Multimodal Benchmarking of Information and Knowledge Algorithms for Data Integration and Decision Intelligence” (Grant number 1OT2OD032581).